\documentclass{article}
\ifdefined\pdfinfoomitdate\pdfinfoomitdate=1\fi
\ifdefined\pdfsuppressptexinfo\pdfsuppressptexinfo=-1\fi

\usepackage{arxiv}
\usepackage[T1]{fontenc}
\usepackage{amsmath}
\usepackage{amssymb}
\usepackage{array}
\usepackage{booktabs}
\usepackage{float}
\usepackage{graphicx}
\usepackage{microtype}
\usepackage{needspace}
\usepackage{placeins}
\usepackage[authoryear,round]{natbib}
\usepackage{xcolor}
\usepackage[colorlinks=true,allcolors=blue!60!black]{hyperref}

\newcommand{\R}{\mathbb{R}}
\newcommand{\E}{\mathbb{E}}
\title{Cross-Prompt Reuse of Entity-Replacement Responses}
\author{
  Jinkun Chen \\
  {\normalfont Dalhousie University} \\
  {\normalfont Halifax, Canada} \\
  {\normalfont\texttt{jinkun.chen@dal.ca}}
  \And
  Xuemin Yu \\
  {\normalfont Dalhousie University} \\
  {\normalfont Halifax, Canada} \\
  {\normalfont\texttt{xuemin.yu@dal.ca}}
  \And
  Sijia Han \\
  {\normalfont Meta} \\
  {\normalfont\texttt{hansijia@meta.com}}
  \And
  Vlado Keselj \\
  {\normalfont Dalhousie University} \\
  {\normalfont Halifax, Canada} \\
  {\normalfont\texttt{vlado@cs.dal.ca}}
}
\hypersetup{pdfauthor={Jinkun Chen, Xuemin Yu, Sijia Han, Vlado Keselj}}

\date{}
\renewcommand{\shorttitle}{Cross-Prompt Response Reuse}
\hypersetup{
  pdftitle={Cross-Prompt Reuse of Entity-Replacement Responses}
}

\makeatletter
\AddToHook{env/table*/begin}{%
  \setlength{\abovecaptionskip}{\@belowcaptionskip}%
  \setlength{\belowcaptionskip}{\@abovecaptionskip}%
}
\makeatother

\hypersetup{bookmarksnumbered=true}

\begin{document}

\maketitle

\begin{abstract}
With model weights fixed, we measure how much of an entity-replacement response
transfers between English factual questions, beyond whether answers change. We add the hidden-state change
from two source questions to a target question about the same entity,
comparing the predicted log-probability shift between two relations' published answers,
given known answer prefixes, with the measured one. Hidden reuse injects the aligned
change at 50\% depth before the target's remaining layers; direct reuse adds source
output-score changes. With other-relation sources, hidden reuse removes 45\% and 49\% of
no-change error on 0.5--1B Qwen and Llama. Gemma-2-2B removes 22\% and
passes one of three comparisons in plain text without its default beginning-of-sequence
(BOS) token. A later protocol restoring it yields 58--68\%, passing every comparison on
three cohorts. Larger checkpoints, chat format and new entities show replication and
format- or cohort-dependent failures. On small plain-format checkpoints, other-relation
hidden reuse mainly reproduces the shift both relations' questions share,
without reliably lowering relation-selective error. With target-relation sources, hidden reuse's
post hoc advantage over direct reuse is 0.2--0.9 nats outside chat format, unconfirmed
and not replacement-specific. The same relation-neutral change at both questions' entity
tokens lowers relation-selective error more than random or other-replacement changes,
indicating receiving-question dependence.
\end{abstract}

% Generated by scripts/paper3_publication_layout.py; edit source sections.
\section{Introduction}

Steering and function vectors reuse recorded activation changes across prompts
\citep{turner2024activationengineering,panickssery2024contrastiveactivation,
todd2024functionvectors}. Accounts of factual recall likewise predict that mid-depth
changes at an entity's tokens can be read out by another question's later layers
\citep{geva2023dissecting,nanda2023factfinding}. A changed answer, however, does not tell
us how much of the receiving question's response has been reproduced. Entity replacement makes that
reference measurable: we can replace the entity in the receiving question itself, which we
call the target, and compare its actual response with the transferred one. How faithfully
can a response recorded in one question reproduce the response in another, and does the
transfer also reproduce the part of that response specific to the relation each question
asks about?

An entity-replacement response is the hidden-state or output-score change caused by
replacing the named entity. We measure its reproduction in answer scores. Replacing SLIB
with JDeveloper in a question about SLIB's developer shifts the log probability of
JDeveloper's published developer answer, \emph{Oracle Corporation}, against its
programming-language answer, \emph{Java}; each answer is scored as a complete string, token
by token after its known preceding tokens. The question about the programming language
scores the same answers in the opposite order. A shift toward \emph{Oracle Corporation} in
both questions is shared; a shift of each question toward its own relation's answer is
relation-selective (Section~\ref{sec:setting}).
Overall fidelity concerns the full measured contrast change.
Each prediction is compared with predicting no change and with transferring a different
replacement's response. Unlike interchange studies, which swap entity representations
between prompts and evaluate the resulting answers
\citep{huang2024ravel,wang2025functionalabstraction}, this reference lets us measure what
share of the target's own signed contrast change a transfer reproduces.

\begin{figure*}[!t]
\centering
\includegraphics[width=\textwidth]{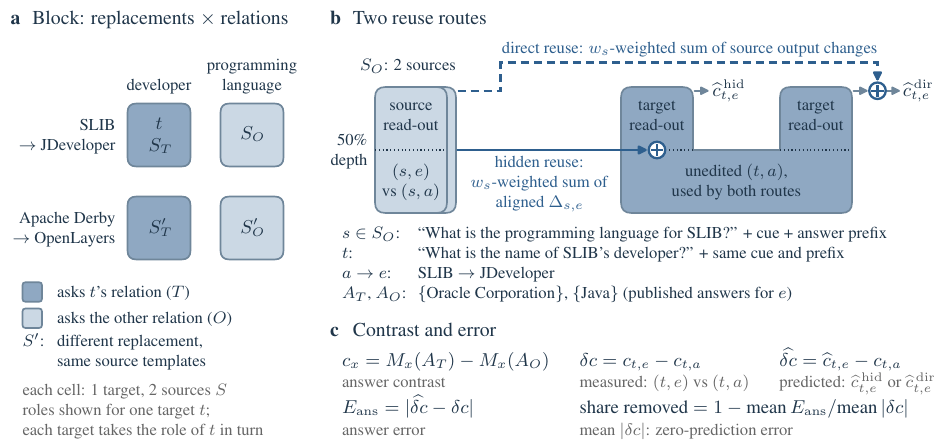}
\caption{Response reuse; model parameters stay fixed. Both routes reuse each source's 50\%-depth change and differ in which read-out, the remaining layers, transforms it: the source's for direct reuse, the target's for hidden reuse. (a)~The three
answer comparisons, pre-specified block sign tests, ask whether hidden reuse from $S_O$ has lower error than each of zero prediction, hidden reuse from $S'_O$ and direct reuse from
$S_O$. In the answer studies, analyses with read-out-matched sources ($S_T$, $S'_T$) are
descriptive, exploratory or post hoc except one failed pre-specified family
(Section~\ref{sec:answer-reproduction}). (b)~In the source and target shapes, depth increases
upward; fills as in~(a). The target's read-out runs twice, with the hidden change (left) and without it (right). $\Delta_{s,e}$ is the
source's 50\%-depth change from $a$ to $e$, recaptured at each known answer prefix; hidden reuse aligns its token rows to target positions
(Equation~\ref{eq:response-memory}). Weights $w_s$ use only unedited states. $\widehat c^{\,\mathrm{hid}}$, $\widehat c^{\,\mathrm{dir}}$: $c$ from each route's predicted scores. The block and example (plain format, cue
\texttt{\textbackslash nAnswer:\textbackslash n}; Appendix~\ref{app:answer-reuse}) are from the earlier crossing cohort, not the 35-block answer cohort, chosen retrospectively for
readable wording, not outcomes (Appendix~\ref{app:crossed-confirmation}). (c)~$M_x(A)$: log
probability mass of group $A$'s complete answers on input~$x$; shares measure contrast
reproduction, not answer correctness or free generation.}
\label{fig:known-edit-overview}
\end{figure*}

We compare two reuse routes that combine the same two source questions with the same
similarity weights (Figure~\ref{fig:known-edit-overview}).
Hidden reuse adds the aligned changes at 50\% depth to the unedited target and continues
its remaining layers; direct reuse adds source output changes. With other-relation sources,
which ask the target's other relation, direct reuse imports that relation's read-out, the
mapping from states to its answers, whereas hidden reuse lets the target's own computation
read the change out. To reproduce the relation-selective part, that computation must then
supply each question's shift toward its own relation's answer, not merely a shift both
share. Target-relation sources ask the target's own relation, so their
read-out matches; we also call them read-out-matched.

The answer to the first question is a portable but incomplete response. In the plain
format, raw text without a chat template or the beginning-of-sequence (BOS) token that
Llama and Gemma add by default, other-relation hidden reuse removes 38--51\% of the
no-change error on Qwen and Llama across the original and two fresh cohorts of new
entities, with format- and cohort-dependent failures. Gemma removes $-$6 to 22\% and passes
at most one of three comparisons per cohort, but 58--69\% with BOS restored, passing every
comparison (Tables~\ref{tab:study-populations} and~\ref{tab:answer-summary}). For the
second question, the relation-selective part is not reliably reproduced on the small
checkpoints in the plain format: hidden reuse there mainly reproduces the shared shift, and
descriptive intervals for selective-error reduction never lie wholly above zero, even where
overall reproduction is substantial. On the original cohort, hidden reuse removes 49\% of
plain-format Llama-3.2-1B's no-change error, yet its selective-error reduction is $-$11.4\%
(Table~\ref{tab:answer-summary}). Pooled over
the three cohorts, target-relation sources narrow hidden reuse's advantage over direct reuse
from 0.55--5.05 to 0.20--0.90 nats, except for plain-format Gemma (post hoc;
Table~\ref{tab:core-matched-routes}). Yet the same relation-neutral change, added to both
questions about a replacement, lowers relation-selective error more than random or
other-replacement changes in pre-specified tests, indicating that these selective effects
depend on the receiving question.

Depth and construction comparisons show where, and in what form, the response remains
reusable. Up to 50\% depth, transferred changes at the entity tokens remove somewhat less
error than the target's own change there (Figure~\ref{fig:answer-depth-exact}).
Span-aligned addition, which aligns the changed entity tokens one to one, outperforms the
best tested single-vector construction on five of six checkpoints, four under pooled
correction across all 223 tests (Table~\ref{tab:testbed}).
Replications span model scales, input formats and new entities within largely familiar
relation/template families.

% Generated by scripts/paper3_publication_layout.py; edit source sections.
\section{Related Work}

\paragraph{Cross-prompt reuse of recorded changes.}
Activation addition and contrastive activation addition transfer prompt-pair differences
\citep{turner2024activationengineering,panickssery2024contrastiveactivation}, and
representation engineering and inference-time intervention shift activations along
directions estimated from many examples \citep{zou2023representation,li2023inference};
function, task, binding and relation vectors transfer recorded states, function vectors
acting when added at early and middle layers
\citep{todd2024functionvectors,hendel2023taskvectors,feng2024binding,wang2024locating}.
Request patching distinguishes receiving-prompt computation at middle layers from
donor-output copying at late layers \citep{variengien2025lookbeforeyouleap}.
States mapped between prompt conditions recover target-prompt behavior
\citep{cheng2026decomposingprompting}; concepts patched across translation prompts are read
out in the receiving language \citep{dumas2024separating}.
Entity representations also transfer across queries. RAVEL interchanges entity
representations between prompts about different attributes or from generic sentences
about the entity \citep{huang2024ravel}, and Patchscopes and related methods decode subject
and other hidden states in other prompts
\citep{ghandeharioun2024patchscopes,pal2023futurelens,chen2024selfie}. REMEDI adds learned
edits to entity representations to change downstream generation
\citep{hernandez2024remedi}. \citet{wang2025functionalabstraction} find that subject-representation interchange,
including from prompts about another relation, largely matches textual substitution in
answer agreement and probabilities, with subject effects in early and middle layers.
Their measures differ from ours, the share of the target's own signed contrast change that
is reproduced, so the two results are not directly comparable;
Section~\ref{sec:testbed} compares state replacement with additive entity-row patches
under our measure.
Knowledge-editing evaluations ask whether an edited fact generalizes to paraphrases
\citep{meng2023memit} and related facts \citep{cohen2024ripple}. We instead measure how much of a replacement's signed
complete-answer change transfers additively across questions and separate shared from
relation-selective components. Mean-centering improves steering vectors \citep{jorgensen2023meancentring}, and steering
evaluations study reliability and context dependence
\citep{tan2024steeringreliability,li2026steeringvectorfields,wu2025axbench}.

\paragraph{Interventions and factual recall.}
Causal mediation studies internal pathways, including gender bias \citep{vig2020causalmediation};
causal tracing localizes factual associations \citep{meng2022locatingeditingfactual},
though localization need not indicate where edits work best \citep{hase2023doeslocalization}.
Causal abstraction formalizes interchange interventions, also on learned subspaces
\citep{geiger2021causalabstractions,geiger2024das,geiger2025causalabstraction}.
Factual recall enriches subjects before extracting attributes \citep{geva2023dissecting}.
Subject lookups do not depend on the preceding context \citep{nanda2023factfinding}, and
both relation-agnostic subject heads and relation-specific components contribute additive
updates \citep{chughtai2024summingfacts,merullo2024language}. Multi-token names may form
implicit vocabulary items \citep{feucht2024tokenerasure}, and heads favoring memorized or
in-context answers compete and can be rescaled
\citep{yu2023characterizing,ortu2024competition}.
\citet{lindsey2025biology} insert California features recorded in another prompt into a
Dallas prompt, eliciting Sacramento; \citet{agarwal2026relation} finds entity-token patches
effective in early and middle layers and final-token patches only in late layers.
These accounts predict the mid-depth transfer whose fidelity we quantify. \citet{zhang2024bestpractices} and \citet{heimersheim2024activationpatching}
discuss patching metrics and interpretation; our equal-length edits follow the former's
symmetric token replacement. Alternative pathways
\citep{makelov2024subspace} and token entrainment \citep{niu2025entrainment} complicate
mechanistic interpretations of agreement.

\paragraph{Perturbations and content controls.}
Interventions can produce off-distribution
representations \citep{grant2025divergent}; random norm-matched vectors can shift
distributions more than steering vectors (\citealp{turner2024activationengineering},
App.~G), and content-free directions move behaviors similarly
\citep{malla2026steeringinterference}, which motivates different-replacement and random
controls.
Complementary work uses stepwise distributional change and uncertainty to diagnose
reasoning failures \citep{chen2026dynamicinstability}; we measure the fidelity of
replacement-induced changes transferred across questions.

% Generated by scripts/paper3_publication_layout.py; edit source sections.
\section{Setting and Response Comparisons}
\label{sec:setting}

\paragraph{Materials and models.}
A relation is the queried attribute, and a template is its wording with an entity slot.
ZsRE, a zero-shot relation extraction dataset, supplies published factual questions,
answers and exact relation labels \citep{levy2017zero}. Each block crosses two entity
replacements with two relations, each with two source questions and one target question;
within each cohort, relation labels, entities, known aliases, templates and complete inputs
are disjoint across blocks, and selection uses factual records, Wikidata entity identities and aliases
\citep{vrandecic2014wikidata} and the tokenizers, before any model outputs were measured. Table~\ref{tab:study-populations} lists the studies,
Appendix~\ref{app:earlier-studies} four earlier next-token studies, and
Appendix~\ref{app:reuse-materials} eligibility, overlap and study order.
The answer results use three cohorts of new-entity blocks whose relation labels and
templates mostly recur from earlier studies (Appendix~\ref{app:answer-confirmation}): the
original cohort of 35 blocks (140 targets) and two fresh cohorts of 33 and 31 blocks;
pooled analyses use all 99 blocks. The earlier 40-block crossing cohort supplies the example of Figure~\ref{fig:known-edit-overview},
earlier next-token tests and the descriptive comparison that motivated other-relation
sources (Appendices~\ref{app:crossed-confirmation}, \ref{app:earlier-studies}
and~\ref{app:study-sequence}).

We use released Qwen2.5-0.5B-Instruct \citep{qwen2025qwen25},
Llama-3.2-1B-Instruct \citep{meta2024llama32} and Gemma-2-2B-it
\citep{gemmateam2024gemma2}, with same-family 7B, 8B and 9B replications.
Changes are injected at an interface, the hidden state at a decoder-layer boundary;
confirmatory interventions use the interface at 50\% depth.
Models run deterministically on CPU in float32; construction and scoring use float64.
Determinism here holds for fixed inputs and batch shapes.
The plain format, fixed by the original protocol before outcomes, omits chat templates and
the default Llama/Gemma BOS token.
Later protocols compare chat format and plain text with only BOS restored (plain+BOS).
The depth scan and the later testbed, read-out, steering-completion and identical-content
studies use each checkpoint's default encoding: plain for Qwen,
whose tokenizer adds no BOS, and plain+BOS for Llama and Gemma.
Appendix~\ref{app:artifacts} gives the exact checkpoints, interfaces and input formats.

\subsection{Response reuse}
Let $H_\ell(x,a)\in\R^{T_x\times D}$ collect hidden states for prompt $x$ with original
entity $a$ (the unedited, or clean, prompt), and let $e$ denote its replacement. A source supplies the finite-difference response
\begin{equation}
\Delta_{s,e}=H_\ell(s,e)-H_\ell(s,a).
\label{eq:source-finite-response}
\end{equation}
Rows (token positions) before the first changed token are zero. In the earlier next-token studies, the matrix
$W^{\mathrm{suffix}}_{s\rightarrow t}$ aligns the first-changed-to-final region by
relative suffix position with linear interpolation, mapping positions between differently
worded prompts. The responses of two sources $S$ yield
\begin{equation}
\widehat\Delta_{t,e}=\sum_{s\in S}w_s
W^{\mathrm{suffix}}_{s\rightarrow t}\Delta_{s,e}.
\label{eq:response-memory}
\end{equation}
The weights are normalized inverse squared distances between unedited final hidden states
(Appendix~\ref{app:transfer-details}). Hidden reuse adds this prediction to the unedited
target states, runs the remaining layers and subtracts the unedited logits
(unnormalized next-token scores).
Direct reuse instead adds the weighted source output changes to the unedited
target scores, using the same sources and weights
(Appendix~\ref{app:direct-output-reuse}). Each source's output change is its own hidden
change continued through the source's remaining layers.
Other-relation sources $S_O$, used in the confirmatory tests, ask the target's other
relation; target-relation sources $S_T$ ask its own relation.
Different-replacement source sets $S'_O$ and $S'_T$ are recorded for the block's other replacement under the same source
templates.

We call this hidden construction suffix interpolation. In answer evaluation it covers the
rendered prompt only, shared candidate-prefix rows align tokenwise, and each candidate
prefix gets its own source capture, the source's change recorded with that prefix
appended, rather than one response reused for the whole answer
(Appendix~\ref{app:answer-reuse}). The answer studies' tests use it; the depth scan and the testbed of standard constructions (Section~\ref{sec:testbed}) also
use span-aligned addition, which copies the changed entity span tokenwise
(Section~\ref{sec:answer-depth}).
Entity replacements preserve token length; source and target questions differ in wording
before the entity.
Internal, unregistered protocols fixed the confirmatory comparisons, and predictions were
saved before
edited-target outcomes (Appendix~\ref{app:answer-confirmation}).

\subsection{Answer contrasts}
Answer evaluation asks whether reuse reproduces the replacement-induced shift
in relative probability between the two relations' complete answers.
For input $x$ rendered for answer completion (Appendix~\ref{app:answer-reuse}), we sum complete-string
probabilities within each answer group and compare their log probability masses.
The fixed groups $A_T$ and $A_O$ contain published answers for the replacement entity
under the target and other relation. In Figure~\ref{fig:known-edit-overview}'s illustrative crossing-cohort developer question, these are \emph{Oracle Corporation} (developer) and \emph{Java} (programming
language), both answers for JDeveloper. We score the same groups before and after
replacing SLIB with JDeveloper.
The group score, answer contrast, replacement-induced change and prediction error are
\begin{equation}
\begin{aligned}
M_x(A)&=\log\sum_{y\in A}p(y\mid x),\\
c_x&=M_x(A_T)-M_x(A_O),\\
\delta c&=c_{t,e}-c_{t,a},\\
E_{\mathrm{ans}}&=\lvert\widehat{\delta c}-\delta c\rvert.
\end{aligned}
\label{eq:answer-response-error}
\end{equation}
Lower error means closer reproduction of the observed signed contrast change;
zero prediction, the no-change baseline, predicts $\widehat{\delta c}=0$, and its mean error
is the no-change error; a predictor's share removed is one minus its mean error
divided by the no-change error. At each
candidate-token position, the recorded
prefix-conditioned response predicts the next-token scores. Full-vocabulary softmax
probabilities yield complete-string scores and a
predicted contrast $\widehat c_{t,e}$, written $\widehat c^{\,\mathrm{hid}}_{t,e}$ or
$\widehat c^{\,\mathrm{dir}}_{t,e}$ for hidden or direct reuse, with
$\widehat{\delta c}=\widehat c_{t,e}-c_{t,a}$. This measures contrast reproduction,
not answer correctness or free generation. Faithfully predicting movement toward
the other relation's answers need not improve factual recall. Qwen 0.5B's original plain-format mean error is 2.869 nats for zero
prediction and 1.572 for hidden reuse (Table~\ref{tab:answer-errors}): 45.2\% of the
no-change error removed. In the original small-checkpoint plain format, the median total
probability mass of the scored candidates ranges from about $10^{-8.4}$ to $10^{-4.6}$
(Table~\ref{tab:answer-candidate-mass}).
Table~\ref{tab:notation} in
Appendix~\ref{app:notation} lists notation and terms.

\Needspace{3\baselineskip}
For each replacement, the two target questions ask different relations, so their answer
groups are swapped: a shift toward the first relation's answers raises the first question's
contrast and lowers the second's. Half the sum of their changes $\delta c_1,\delta c_2$
keeps each question's movement toward its own relation's answers and cancels a shift shared
by both questions; half the difference keeps that shared shift:
\begin{equation}
s=\frac{\delta c_1+\delta c_2}{2},
\qquad
h=\frac{\delta c_1-\delta c_2}{2},
\label{eq:answer-paired-components}
\end{equation}
the relation-selective component $s$ and the shared component $h$, oriented by the block's
first relation (the subscript $s$ in
Equations~\ref{eq:source-finite-response} and~\ref{eq:response-memory} instead indexes
sources). For predicted $\widehat s$ and $\widehat h$, selective error is
$\lvert\widehat s-s\rvert$ and shared error $\lvert\widehat h-h\rvert$; selectivity tests ask
whether $\lvert s\rvert-\lvert\widehat s-s\rvert$, the reduction from zero prediction, is
positive.

\paragraph{Statistical analysis.}
Paired effects, baseline error minus evaluated-prediction error, average the four targets
of each block. Exact sign tests, one-sided unless stated, count blocks with positive and negative effects,
assessing how broadly improvement recurs; medians of block effects get distribution-free
intervals by inverting the sign test, and means with pointwise bootstrap intervals describe
magnitude. Holm correction \citep{holm1979simple} adjusts within each pre-specified family
at 0.05; Table~\ref{tab:study-populations} gives each study's tests and passes, and passes
in the text are within-family. Pre-specified (confirmatory) comparisons were fixed in a
written protocol before their outcomes, and one passes when its Holm-adjusted $p$ is below
0.05; descriptive results report estimates without tests, and post hoc analyses were
chosen after outcomes were seen. With 31--35 nonzero blocks, 80\% power requires a true positive-sign probability of
about 71--80\%, depending on family size and Holm rank
(Appendix~\ref{app:statistical-details}); a failed test does not show equivalence. Pooling all 223 confirmatory tests into one family
(pooled correction) would change 26 conclusions (Table~\ref{tab:round10-pooled}). Intervals resample whole blocks 20,000 times
(Appendix~\ref{app:evidence-fidelity}). Cross-cohort tests assume block independence despite
reused relation labels. Label-resampled intervals, which resample relation labels, are post hoc sensitivity summaries that widen even for
independent blocks, not dependence-adjusted tests; Holm addresses multiplicity, not this dependence
(Appendix~\ref{app:statistical-details}).
Appendices~\ref{app:evidence-fidelity}--\ref{app:evidence-content} give the answer
families and post hoc checks. The answer depth and position scans are descriptive.

% Generated by scripts/paper3_round12_tables.py; do not edit by hand.
\begin{table*}[tbp]
\centering
\footnotesize
\caption{Studies and their pre-specified test families. Tests: number of tests; Pass: passes under Holm correction within each family; Lost: passes lost when all 223 confirmatory tests form one Holm family (Table~\ref{tab:round10-pooled}). Each family was fixed by an internal written protocol before its outcomes, although for Qwen some testbed arms include blocks already observed (Appendix~\ref{app:study-sequence}). Material: blocks new to the study, reused from earlier studies, or both (earlier: these studies' own materials, which precede the answer studies). Blocks hold four targets (components for two earlier next-token studies; Table~\ref{tab:direction-studies}). Protocols named one family primary without making the others conditional on it. Two steering-completion tests repeat testbed tests (Appendix~\ref{app:evidence-constructions}). BOS: beginning-of-sequence token.}
\label{tab:study-populations}
\setlength{\tabcolsep}{3pt}
\begin{tabular}{@{}lllrrrl@{}}
\toprule
Study & Material & Blocks & Tests & Pass & Lost & Main result \\
\midrule
Answer confirmation & new & 35 & 9 & 7 & 3 & 45/49/22\% of no-change error removed \\
Larger checkpoints & reused & 35 & 9 & 7 & 0 & 51/46/14\% removed \\
Chat format & reused & 35 & 18 & 18 & 4 & All comparisons pass, Gemma's included \\
First fresh cohort & new & 33 & 20 & 14 & 4 & 7--8B selectivity tests pass in family; Gemma passes none \\
BOS token & both & 99 & 18 & 18 & 0 & Gemma passes every comparison \\
Second fresh cohort & new & 31 & 38 & 14 & 3 & Matched-route test fails on all six \\
Answer depth and rows & reused & 35 & \multicolumn{3}{c}{descriptive} & Share declines from the input \\
Testbed and read-out & reused & 64--99 & 30 & 20 & 3 & Single vectors trail span alignment \\
Steering completion & reused & 64--99 & 24 & 13 & 4 & Best vector trails span alignment on five of six \\
Identical-content controls & reused & 99 & 12 & 12 & 2 & Neutral content beats random and other-replacement changes \\
Earlier next-token studies & earlier & 10--45 & 45 & 36 & 3 & Same beats different replacement \\
\bottomrule
\end{tabular}
\end{table*}

% Generated by scripts/paper3_publication_layout.py; edit source sections.
\section{Cross-Prompt Response Reuse}
\label{sec:response-memory}

We first ask how much of the target's response transfers, against which comparators, and
whether its relation-selective part is reproduced (Section~\ref{sec:answer-reproduction});
we then ask where in depth and token position, and in which constructions, the changes
remain reusable (Sections~\ref{sec:answer-depth} and~\ref{sec:testbed}).

\subsection{Answer-conditioned response reproduction}
\label{sec:answer-reproduction}
On 140 new-entity targets in 35 disjoint blocks of four, absolute answer-contrast error
(Equation~\ref{eq:answer-response-error}) is averaged within blocks.
Confirmatory sources ask about the other relation, a condition chosen after a descriptive
analysis of the crossing cohort in which it separated the routes most clearly, whereas the
route-gap intervals with target-relation sources included zero on Llama and Gemma
(Appendix~\ref{app:study-sequence}). Pre-specified families compare, on each checkpoint,
same-replacement hidden reuse from $S_O$ with direct reuse from $S_O$, zero prediction and
different-replacement hidden reuse from $S'_O$, which checks that predictions depend on the
replacement.

% Generated by scripts/paper3_round11_tables.py; do not edit by hand.
\begin{table}[!t]
\centering
\footnotesize
\caption{Answer reuse at 1/2 depth. Overall (other-relation sources): \% of the no-change error removed by same-replacement hidden reuse in each cohort (35/33/31 blocks; chat: original blocks), with passes of the three pre-specified comparisons in parentheses (--: untested). Gap (target-relation sources): direct minus hidden error, nats, over 99 blocks (chat: 35); positive favors hidden reuse (Tables~\ref{tab:core-matched-routes} and~\ref{tab:core-extra-comparisons}b). Sel. (other-relation sources): selective-error reduction from zero prediction (\%), original 35 blocks (Table~\ref{tab:selectivity-checks}). Gap and Sel. are descriptive. Robust shares: Table~\ref{tab:round12-robust}.}
\label{tab:answer-summary}
\setlength{\tabcolsep}{1.5pt}
\begin{tabular}{@{}llrrr@{\hspace{4.3pt}}r@{\hspace{4.3pt}}r@{}}
\toprule
\multicolumn{2}{@{}l}{Source relation:} & \multicolumn{3}{c}{Other} & Target & Other \\
\cmidrule(lr){3-5}\cmidrule(lr){6-6}\cmidrule(l){7-7}
\multicolumn{2}{@{}l}{Metric:} & \multicolumn{3}{c}{Overall \% (passes)} & Gap/nats & Sel. \% \\
Model & Input & Orig. & Fresh 1 & Fresh 2 & \shortstack{$n=99$\\chat: 35} & \shortstack{Orig.\\$n=35$} \\
\midrule
Qwen 0.5B & plain & 45 (3) & 42 (3) & 38 (2) & 0.32 & 8.7 \\
 & chat & 43 (3) & -- & -- & 0.63 & 21.8 \\
\addlinespace
Llama 1B & plain & 49 (3) & 50 (3) & 40 (1) & 0.24 & $-$11.4 \\
 & +BOS & 57 (--) & 66 (--) & 57 (--) & 0.20 & 8.3 \\
 & chat & 56 (3) & -- & -- & 0.37 & 40.7 \\
\addlinespace
Gemma 2B & plain & 22 (1) & 2 (0) & 18 (1) & 0.29 & $-$9.8 \\
 & +BOS & 68 (3) & 65 (3) & 58 (3) & 0.26 & 35.9 \\
 & chat & 46 (3) & -- & -- & 0.34 & 38.3 \\
\addlinespace
Qwen 7B & plain & 51 (3) & 42 (3) & 49 (3) & 0.90 & 50.2 \\
 & chat & 35 (3) & -- & -- & 1.20 & 37.2 \\
\addlinespace
Llama 8B & plain & 46 (3) & 49 (3) & 50 (3) & 0.40 & 19.7 \\
 & +BOS & 41 (--) & 51 (--) & 54 (--) & 0.36 & 2.6 \\
 & chat & 20 (3) & -- & -- & 0.21 & 6.7 \\
\addlinespace
Gemma 9B & plain & 14 (1) & $-$6 (0) & 9 (1) & 0.45 & $-$7.3 \\
 & +BOS & 65 (3) & 69 (3) & 66 (3) & 0.58 & 48.7 \\
 & chat & 52 (3) & -- & -- & 0.20 & 52.1 \\
\bottomrule
\end{tabular}
\end{table}

\paragraph{How much transfers.}
In the plain format on the original cohort, hidden reuse removes 45--51\% of the
no-change error on Qwen and Llama but 14--22\% on Gemma (Table~\ref{tab:answer-summary}).
Qwen and Llama pass all three comparisons at both scales, whereas Gemma passes one of three
at each scale. With the BOS token restored, both Gemma checkpoints remove 58--69\% across
the three cohorts and pass every comparison; the token also enlarges the observed changes,
so shares are not directly comparable across formats
(Tables~\ref{tab:round10-families} and~\ref{tab:round10-bos}). In each checkpoint's chat
template all eighteen comparisons pass (Table~\ref{tab:round9-families}). On two fresh
cohorts with new entities, Qwen2.5-7B and Llama-3.1-8B pass every comparison;
the small checkpoints pass none against zero prediction on the second in the plain format,
and plain-format
Gemma at most one per cohort (Tables~\ref{tab:round9-families}
and~\ref{tab:round10-families}).

The shares are ratios of means of right-skewed block gains; over the 99 blocks the median
block loses 32--51\% of its no-change error on Qwen and Llama in the plain format and
61--64\% on Gemma with the BOS token (Table~\ref{tab:round12-robust}). Excluding targets
whose answers share a word with an entity name lowers plain-format Gemma's pooled shares to
near zero but changes the other pooled shares by at most five points
(Table~\ref{tab:round12-robust}). Without the blocks containing those targets, every Qwen
and Llama comparison in the original plain format still passes (post hoc;
Table~\ref{tab:core-extra-comparisons}a), but plain-format Gemma-2-2B's sole original pass,
against direct reuse, no longer does ($p_{\mathrm{Holm}}=0.025$ to $0.078$;
Tables~\ref{tab:answer-confirmation-tests} and~\ref{tab:core-extra-comparisons}a).

At the answer cue, before any candidate token is scored, hidden reuse reduces the
full-vocabulary Kullback--Leibler divergence from the edited target by
88--97\% relative to the unedited target in each checkpoint's default encoding, against
44--87\% for other-relation direct reuse (Table~\ref{tab:round11-readouts}). In greedy
generation, which follows its own tokens, one question-plus-cue capture (one source recording at
the question and cue, without answer tokens) with
span-aligned addition from other-relation sources reproduces the edited answer string for 5--23\% of targets within 16
tokens (Table~\ref{tab:round11-generation}); construction, capture and scope differ from
candidate scoring, so the gap does not isolate autoregressive error accumulation.

\paragraph{Against which comparators.}
How much hidden reuse gains over direct reuse depends on which relation the sources ask.
With other-relation sources, direct reuse imports that relation's read-out: in the plain
format it removes almost none of the no-change error on the small Qwen and Gemma
checkpoints and adds 12--28\% to it at 7--9B; in the chat format it adds 63--78\% at 7--9B
(Tables~\ref{tab:answer-errors} and~\ref{tab:core-extra-comparisons}b). With the three
cohorts pooled, using target-relation instead of other-relation sources narrows the
hidden-over-direct gap from 0.55--5.05 to 0.20--0.90 nats, except for plain-format Gemma
(Table~\ref{tab:core-matched-routes}); this switch changes the recorded response as well
as its read-out.
Target-relation direct reuse can also outperform other-relation hidden reuse:
on the original blocks in chat format it does so on Gemma-2-2B, Llama-3.1-8B and
Gemma-2-9B, by up to 3.08 nats with pointwise intervals excluding zero
(Table~\ref{tab:core-extra-comparisons}b).

With target-relation sources, the residual advantage of hidden reuse is unconfirmed. It is
0.24--0.90 nats pooled over the three plain-format cohorts and 0.20--0.58 nats with the BOS
token (Table~\ref{tab:core-matched-routes}; post hoc). In chat format on the original
blocks it is 0.20--1.20 nats, with pointwise intervals including zero on Gemma-2-2B,
Llama-3.1-8B and Gemma-2-9B (Table~\ref{tab:core-extra-comparisons}b; post hoc). On three
plain-format checkpoints,
the label-resampled interval for the mean gap includes zero or the interval for the share
of favoring blocks includes one half (Table~\ref{tab:core-matched-routes}). After each
route is rescaled, the advantage is 0.11--0.75 nats in the plain format on the fresh
cohorts; Gemma-2-2B's interval includes zero (Table~\ref{tab:round12-routes}).
A pre-specified test on the 31-block second cohort
failed on all six checkpoints
(Table~\ref{tab:round10-neutral-route}). Nor is the route advantage
replacement-specific: with target-relation sources, a different replacement gains more from
the hidden route than the correct one on every checkpoint in the plain and plain+BOS
formats, by 0.24--1.72 nats over the 99 blocks with intervals excluding zero in eight of the ten
cases, consistent with the target attenuating
wrong injected changes (Table~\ref{tab:round12-routes}).

% Generated by scripts/paper3_round11_runs_tables.py; do not edit by hand.
\begin{table}[tbp]
\centering
\footnotesize
\caption{Best single-vector construction at the 1/2 interface, default encoding. The best of 90 single-vector constructions is selected on the original cohort; Best and Span (span-aligned addition) are evaluated on the same 64 already-studied new-entity blocks. Best/Span: zero-prediction error removed (\%). Gain: mean vector minus span error (nats), with 95\% label-resampled intervals. $^*$: passes its pre-specified six-test Holm family (Appendix~\ref{app:evidence-constructions}); $^\dagger$: fails the 223-test pooled correction. The Qwen 0.5B and Gemma 9B comparisons repeat earlier tests on the same blocks (Section~\ref{sec:testbed}).}
\label{tab:testbed}
\setlength{\tabcolsep}{3pt}
\begin{tabular}{@{}lrrr@{}}
\toprule
Checkpoint & Best & Span & Gain [95\% interval] \\
\midrule
Qwen 0.5B & 13.3 & 47.3 & 0.85$^{*}$ [0.33, 1.62] \\
Llama 1B & 38.3 & 65.3 & 0.66$^{*}$ [0.35, 1.05] \\
Gemma 2B & 42.1 & 65.2 & 0.82$^{*}$ [0.41, 1.25] \\
Qwen 7B & 29.9 & 48.5 & 1.38$^{*\dagger}$ [0.18, 2.50] \\
Llama 8B & 45.1 & 55.5 & 0.48 [$-$0.11, 1.10] \\
Gemma 9B & 48.1 & 69.8 & 1.09$^{*}$ [0.50, 1.77] \\
\bottomrule
\end{tabular}
\end{table}

\paragraph{What transfers.}
On the original plain-format blocks, the selective component $s$ carries 9--38\% of the
observed variance, less on the small checkpoints than at 7--9B; the rest is shared between
the questions (Table~\ref{tab:core-component-variance}). Hidden reuse cuts shared error by
34--62\% on the small checkpoints (Table~\ref{tab:answer-components}). Descriptive
pointwise 95\% intervals for selective-error reduction are never wholly positive on the
small plain-format checkpoints in any cohort; on the original cohort they are positive for
plain Qwen2.5-7B and Llama-3.1-8B, Gemma with BOS, and every chat checkpoint except
Llama-3.1-8B (Table~\ref{tab:selectivity-checks}). The pre-specified 7--8B selectivity
tests followed up these descriptive reductions: both passed on the first fresh cohort,
neither under pooled correction, and only Llama-3.1-8B passed on the second
(Table~\ref{tab:selectivity-checks}). A
constant push toward each question's own answers, predicting the same small positive
$\widehat s$ for every pair, would pass all four tests, since most observed $s$ are positive. On these two checkpoints in the fresh cohorts, same-replacement
hidden reuse lowers selective error by 34--36\%, whereas different-replacement hidden reuse
lowers it by only $-$9 to 5\% (Table~\ref{tab:selectivity-checks}).

Relation-neutral sources, ``Tell me about $X$.''\ and ``What
do you know about $X$?'', continued through the target, remove 39--45\% of the no-change error
on Qwen2.5-7B and Llama-3.1-8B, versus 42--51\% for other-relation sources
across the three cohorts (Appendix~\ref{app:neutral-cohort-details}).
The post hoc advantage of other-relation over neutral
sources, 0.24--0.45 nats on plain-format Qwen and Llama pooled over cohorts, keeps a mean interval above zero under
resampled labels only at 0.5--1B and an interval for the share of favoring blocks above
one half only on Qwen2.5-0.5B
(Table~\ref{tab:core-local-sensitivity}a). The six pre-specified comparisons
on the 31-block second cohort did not detect this advantage
(Table~\ref{tab:round10-neutral-route}).

Injected with equal weights into both relation-swapped targets, the relation-neutral
entity-row change is identical for the two questions, so any shift of each question toward
its own relation's answers must come from the receiving question. This identical change
lowers relation-selective error in 60--83 of 99 blocks, passing all six pre-specified tests
within their family (Table~\ref{tab:round11-families}). Identical random
perturbations with the same row norms, and the other replacement's neutral change given
identically to both targets, lower selective error by at most 0.22 nats; the neutral change
beats both on all six checkpoints in pre-specified families
(Tables~\ref{tab:controls-families}--\ref{tab:controls-reductions}).
For all three comparisons, Llama-3.2-1B's within-family pass does not survive pooled
correction. Under label resampling, some Llama-3.2-1B mean-effect intervals include zero,
and several favoring-block intervals include one half
(Tables~\ref{tab:core-local-sensitivity}b and~\ref{tab:controls-intervals}). These tests use the
same selective-error readout as the selectivity checks above but inject identical content
into both questions; they also differ in construction, encoding (each checkpoint's
default, with BOS for Llama and Gemma) and inference (sign tests over 99 blocks rather
than per-cohort intervals).

Blocking attention from the answer positions to entity rows above the interface
\citep{geva2023dissecting} retains only 3--15\% of the error reduction from transferred
entity rows on Qwen and Llama but more than half on Gemma; blocking the other prompt rows
except the first token and the final cue row retains nearly all of Llama's reduction, about
half of Gemma's and little of Qwen's. The pre-specified test that blocking entity rows raises block error more than
blocking the other rows passes only on Llama (Tables~\ref{tab:round11-families}
and~\ref{tab:round11-knockouts}).

\begin{figure*}[!t]
\centering
\includegraphics[width=\textwidth]{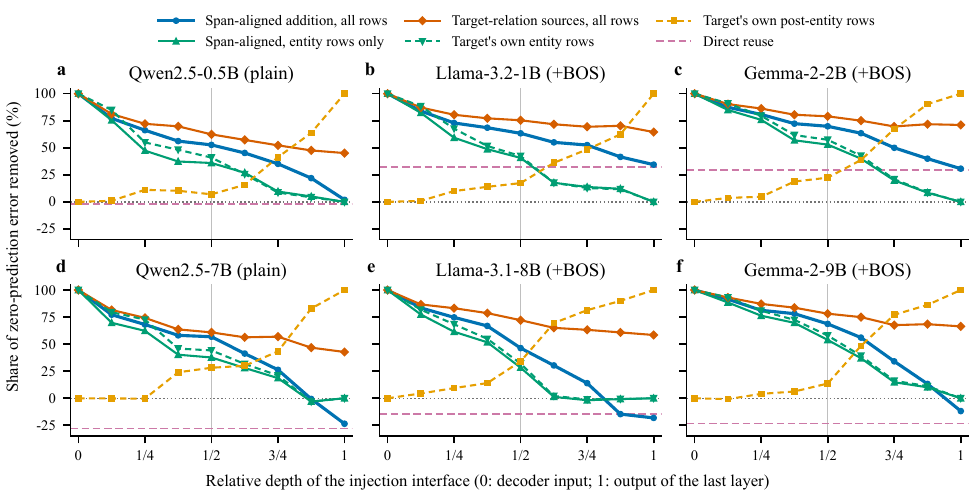}
\caption{Answer-error reduction relative to zero prediction over 35 blocks, default
encoding. Solid: span-aligned addition, with other-relation sources unless labelled
target-relation; dashed: target-own edited-minus-clean patches, except the horizontal line,
other-relation direct reuse; the legend gives row coverage. Vertical: the confirmatory
interface; the curves use span-aligned addition, not the suffix interpolation of
Table~\ref{tab:answer-summary}.
The terminal interface $d=1$ precedes final normalization.
Appendix~\ref{app:evidence-depth} gives numerical summaries and row definitions,
and explains why Qwen2.5-7B uses its saved plain-format scan.}
\label{fig:answer-depth-exact}
\end{figure*}

\subsection{Depth and row position of answer reuse}
\label{sec:answer-depth}

We next ask where in the target computation these responses remain reusable.
A pre-specified descriptive scan injects the same responses at nine interfaces from
the decoder input ($d=0$) to the last layer ($d=1$), keeping the source weights from 50\%
depth. Span-aligned addition copies the changed entity span tokenwise, interpolates only
the later question and cue rows, and copies answer-prefix rows. At each replacement
endpoint, changed-span token IDs match across source and target questions, so at $d=0$
it reproduces the edited target by construction. Row subsets separate entity rows from post-entity question, cue and answer-prefix rows. Oracle patches add the target's own
edited-minus-clean states to the same rows (Appendix~\ref{app:evidence-depth}).

In each checkpoint's default encoding the share removed by span-aligned addition declines
with depth on every checkpoint, from 100\% at $d=0$ to 46--70\% at $d=1/2$
(Figure~\ref{fig:answer-depth-exact}). Up to 50\% depth, transferred entity rows remove
somewhat less error than the target's own entity-row patch
(Table~\ref{tab:round11-depth-default}).
With target-relation sources, all aligned rows remove more error than entity rows alone at
$d=1/2$, with pointwise intervals excluding zero on all five checkpoints with these runs
(Table~\ref{tab:round11-rows}). At $d=7/8$ the target's own post-entity change removes
most of the no-change error, but other-relation responses on those rows remove far less
(Table~\ref{tab:round11-depth-default}). Row effects are not additive: at $d=1/2$, the
target's own entity-row and post-entity-row changes, continued separately and added as
predicted contrast changes, remove 49--69\% of the no-change error, whereas patching both
together reproduces the edit
(Appendix~\ref{app:row-additivity}). Omitting answer-prefix rows keeps most of the
span-aligned share at $d=1/2$
(Tables~\ref{tab:round11-depth-default} and~\ref{tab:round11-rows}), so under candidate
scoring one question-plus-cue capture carries most of this gain.

\subsection{Standard constructions on the testbed}
\label{sec:testbed}

Using the same answer-error reference, we ask whether a single vector can substitute
for the aligned change matrix. The testbed protocol compares standard constructions over
the 99 blocks at the 1/2 interface in default encoding. Its simplest
single vector is the weighted source change at the last entity row, kept at its native
scale and added at every target row from the entity onward; contrastive activation
addition instead starts after the prompt \citep{panickssery2024contrastiveactivation}.
This broadcast vector has larger mean error than zero prediction on all six checkpoints
(Table~\ref{tab:core-extra-comparisons}c), and span-aligned addition beats it in 95--99
blocks (Table~\ref{tab:round11-families}). At matched entity-row coverage, span-aligned
changes lower error by 0.30--0.49 nats relative to a native-scale vector
(descriptive; Table~\ref{tab:core-extra-comparisons}c).
Entity-state replacement \citep{huang2024ravel,wang2025functionalabstraction} and additive
entity-row patches differ by $-$0.06 to 0.01 nats, undistinguished by the pre-specified tests.
Span-aligned addition also lowers mean error relative to the confirmatory suffix
interpolation by 0.12--0.28 nats, with all six pointwise intervals excluding zero
(descriptive; Table~\ref{tab:core-extra-comparisons}c).

Two protocols then tuned single vectors on the original cohort and tested them on the two
fresh cohorts, which earlier studies had already used (64 blocks). The testbed's tuned
comparison (D1c in Table~\ref{tab:round11-families}) selects the coefficient of one
last-entity-row vector added at each entity row, without broadcasting it to later rows.
The later steering-completion protocol of Table~\ref{tab:study-populations} selects the
best of 90 constructions by minimizing mean block answer error on the original cohort; they
vary averaging (single-pair or sixteen-context), direction (entity or answer cue),
placement and amplitude (Table~\ref{tab:testbed}). After tuning, the broadcast vector's pre-specified test
against zero passes on three checkpoints, one under pooled
correction. Rowwise norm matching scales one direction to the span-aligned row norms and
beats zero on five checkpoints over all 99 blocks, four under pooled
correction. Sixteen-context averaging gives no detected gain in its pre-specified comparison.
Span-aligned addition beats the best of the 90 constructions (a norm-matched direction on
four checkpoints) on all but Llama-3.1-8B under family-wise correction and on four under
pooled correction; two of the five passes repeat earlier tests
(Appendix~\ref{app:evidence-constructions}).

\FloatBarrier
% Generated by scripts/paper3_publication_layout.py; edit source sections.
\section{Discussion}
\label{sec:discussion}

\paragraph{What the quantitative profile adds.}
For reuse of recorded changes, overall fidelity and relation selectivity can diverge:
measured against the target's own response, overall fidelity is substantial but
incomplete, whereas relation selectivity is not reliably reproduced on the small
plain-format checkpoints, where the transfer mainly reproduces the shift both questions share. Binary answer outcomes do not quantify
the magnitude of the reproduced response or its shared and relation-selective components.
Beyond established entity-state transfer
\citep{geva2023dissecting,nanda2023factfinding}, target-relation sources substantially
narrow the hidden-over-direct gap except for plain-format Gemma, and the identical
relation-neutral change retains selective effects that depend on the receiving question,
beyond random and other-replacement controls.

\paragraph{Which changes to inject.}
Aligned changes on all rows beat entity rows alone at 50\% depth
(Section~\ref{sec:answer-depth}) and, on five of six checkpoints, the best tested single
vectors (Section~\ref{sec:testbed}). Scaling single vectors to span-aligned row norms narrows that
gap, but directions, coverage and prefix information also vary, so row structure is not
isolated.

% Generated by scripts/paper3_publication_layout.py; edit source sections.
\section{Conclusion}
\label{sec:conclusion}

With known answer prefixes supplied, entity-replacement responses are partly portable
across questions: other-relation hidden
reuse removes 38--51\% of the no-change error on plain-format Qwen and Llama and 58--69\%
on Gemma with BOS restored, with format- and cohort-dependent exceptions. On the
small checkpoints in the original plain format, other-relation hidden reuse mainly
reproduces the shared shift
without reliably lowering relation-selective error. The
identical relation-neutral change can nevertheless produce
selective effects that depend on the receiving question. With the three cohorts pooled,
target-relation sources substantially narrow the hidden-over-direct advantage except for
plain-format Gemma; the residual advantage is unconfirmed and not replacement-specific.
Span-aligned addition beats the best tested single vectors on five of six checkpoints, four
under pooled correction. Together, these measurements quantify the magnitude, relation
selectivity and construction dependence of cross-question response reuse.

% Generated by scripts/paper3_publication_layout.py; edit source sections.
\section{Limitations}

Relation labels and question wording vary together, and some labels express overlapping
predicates, so they do not isolate semantic relation from wording. The fresh cohorts test
new replacements within largely familiar relation and template families
(Appendix~\ref{app:reuse-materials}). Eligibility requires equal-length token-span
replacements in English questions or completion prompts. Results cover six instruct
checkpoints from three families, with size and training lineage confounded.
Chat jointly changes the template, BOS token, answer instruction and
scored terminator (end-of-turn token instead of newline). Restoring BOS alone reproduces
Gemma's chat-format improvement but not Llama-3.2-1B's chat-format selectivity gain, whose source among the
other components is unresolved. The three answer comparisons were not tested for Llama
with BOS, and fresh cohorts were not run in chat format.

Confirmatory studies use one interface at 50\% depth. The descriptive answer depth and
position scans reuse the observed 35-block cohort and split constructions by target
rows; their target-relation row subsets and neutral-source runs lack Qwen2.5-7B, whose
re-run failed its consistency check (Appendix~\ref{app:evidence-depth}). Suffix interpolation is
inexact at the decoder input and lets entity rows receive non-entity source rows.
Span-aligned addition applies only to same-replacement sources; different-replacement and
random controls use interpolation, except in the identical-content experiment, whose
other-replacement control interpolates within the entity span
(Appendix~\ref{app:evidence-content}). Attention knockouts locate attention paths rather
than heads or features and also shift the unedited answer contrast. Construction comparisons
jointly vary coverage, direction and prefix information and do not evaluate behavioral
steering; norm-matched vectors borrow span-aligned norms
(Appendix~\ref{app:evidence-constructions}).

Complete-answer evaluation conditions sources and targets on known published-answer
prefixes. Its literal candidate groups carry little absolute probability mass and
exclude alternative aliases and formatting, including semantically equivalent strings
\citep{holtzman2021surface}; Appendix~\ref{app:answer-candidate-mass} gives their masses.
The next-token distribution at the answer cue and greedy generation from a single capture
are separate measures: one question-plus-cue capture with other-relation sources reproduced the
edited greedy answer string for only 5--23\% of targets
(Appendix~\ref{app:evidence-depth}).
Component splits are algebraic output contrasts, not identified hidden-state mechanisms.

The protocols were internal, not externally registered, and fixed confirmatory
comparisons before their own outcomes. Later designs used earlier findings and
materials, including the choice of other-relation sources, the BOS
ablations after Gemma's plain-format failures, and the 90-construction comparison, whose
distinct selection and evaluation cohorts had all been studied before. Appendix~\ref{app:study-sequence} gives the order and
scope of each protocol. The neutral-source controls use two fixed English templates, and the
different-replacement control changes both the original and replacement entity.

Sign inference assumes independent blocks or relation/template components. Grouping
recorded overlaps leaves possible dependence through aliases, recurring answers and
relation labels reused across cohorts. Label resampling shows sensitivity for some mean
effects and positive-block shares but gives descriptive intervals, not
dependence-adjusted family-wise tests (Appendix~\ref{app:statistical-details}).
Study-wide Holm correction addresses a separate multiplicity question and changes
26 conclusions, including first-fresh-cohort selectivity and some tuned-vector and
identical-content results; Llama-3.1-8B's second-cohort selectivity replication remains
(Table~\ref{tab:round10-pooled}). Limited power also precludes interpreting failed
comparisons as equivalence (Appendix~\ref{app:statistical-details}).
A constant push would also pass the fresh-cohort selectivity sign tests, so interpretation
relies on magnitudes and different-replacement comparisons (Table~\ref{tab:selectivity-checks}).

\begingroup
\raggedright

\endgroup

\clearpage
\appendix
\section*{Appendix guide}
Appendix~\ref{app:transfer-details} defines the constructions and scores and collects
notation in Table~\ref{tab:notation}. Appendix~\ref{app:reuse-materials} describes the
blocks, cohorts, checkpoints and formats, and the order of the studies
(Appendix~\ref{app:study-sequence}). The remaining appendices follow
Section~\ref{sec:response-memory}: Appendices~\ref{app:evidence-fidelity} (prediction
fidelity, calibration and statistical procedures) and~\ref{app:evidence-replication}
(scale, format, fresh-cohort and source-relation comparisons) support
Section~\ref{sec:answer-reproduction}; Appendix~\ref{app:evidence-depth} supports
Section~\ref{sec:answer-depth} and the cue-distribution and generation results of
Section~\ref{sec:answer-reproduction}; Appendix~\ref{app:evidence-constructions} supports
Section~\ref{sec:testbed}; and Appendix~\ref{app:evidence-content} gives the attention
knockouts and identical-content controls of Section~\ref{sec:answer-reproduction}.
Table~\ref{tab:study-populations} inventories every test family, and
Table~\ref{tab:round10-pooled} lists the conclusions that change under pooled correction.
% Generated by scripts/paper3_publication_layout.py; edit source sections.
\section{Response Construction and Scores}
\label{app:transfer-details}

\paragraph{Notation and terms.}
\label{app:notation}
Table~\ref{tab:notation} collects the terms. Appendix~\ref{app:earlier-studies}
defines the earlier next-token direction comparisons.

\begin{table*}[!htbp]
\centering
\small
\caption{Notation and terms. Source sets $S_T$ and $S_O$ ask about the target's relation
and the other relation; primed sets ($S'_T$, $S'_O$) are recorded for the block's other
replacement under the same source templates. $\widehat c^{\,\mathrm{hid}}_{t,e}$ and
$\widehat c^{\,\mathrm{dir}}_{t,e}$ are the contrasts predicted by hidden and direct reuse.}
\label{tab:notation}
\begin{tabular}{@{}>{\raggedright\arraybackslash}p{0.27\linewidth}>{\raggedright\arraybackslash}p{0.68\linewidth}@{}}
\toprule
Term & Meaning \\
\midrule
Replacement $a\to e$ & Swapping the original entity $a$ for an entity $e$ with the same token length in a question \\
Replacement response & Hidden-state or output-score difference between the replaced and original question (Equation~\ref{eq:source-finite-response}) \\
Source, target & Questions whose recorded responses are reused; the question whose response is predicted \\
Hidden reuse & Aligned, weighted source hidden responses added at 50\% depth and continued through the target's remaining layers \\
Direct (output) reuse & Weighted source output-score changes added to the target's unedited scores \\
Zero prediction & Predicts no change; its answer error, the zero or no-change error, is the size of the observed change \\
Same, different replacement & Sources recorded for the target's own replacement, or for the block's other replacement under the same source templates \\
Target-, other-relation sources & Sources that ask about the target's relation, or about the block's other relation \\
Native amplitude; row-norm profile & Measured source magnitudes; a prescribed vector length at each token used to match magnitudes \\
Donor, reference & In magnitude matching, the response being rescaled and the response supplying row norms \\
Block; component & Four target cells forming one inference unit; relations linked by shared templates \\
KL recovery $R_{\mathrm{KL}}$ & One minus the prediction's Kullback--Leibler error relative to the unedited target's (Equation~\ref{eq:kl-recovery}) \\
$c$, $\delta c$, $E_{\mathrm{ans}}$ & Answer contrast, its replacement-induced change and the absolute prediction error (Equation~\ref{eq:answer-response-error}) \\
$s$, $h$; $\widehat s$ & Selective and shared components of a relation-swapped pair (Equation~\ref{eq:answer-paired-components}); a prediction's selective component, with selective error $\lvert\widehat s-s\rvert$. The subscript $s$ of Equations~\ref{eq:source-finite-response} and~\ref{eq:response-memory} instead indexes source questions \\
Share removed & One minus a predictor's mean answer error divided by the zero-prediction error \\
\bottomrule
\end{tabular}
\end{table*}

\paragraph{Suffix correspondence and source weights.}
Let $p_x$ be the first changed position and $T_x$ the sequence length. For target
row $j\ge p_t$, the source coordinate is
\begin{equation}
x_{s\rightarrow t}(j)=p_s+(T_s-1-p_s)
\frac{j-p_t}{T_t-1-p_t}.
\label{eq:suffix-coordinate}
\end{equation}
Writing $q=\lfloor x\rfloor$ and $\alpha=x-q$, the aligned row is
$(1-\alpha)\Delta_s[q]+\alpha\Delta_s[\min(q+1,T_s-1)]$.
Rows before $p_t$ are zero. The map connects the first changed and final rows;
it is a positional heuristic rather than a claim of semantic correspondence.
For normalized unedited final states
$z_x=H_\ell(x,a)_{T_x-1}/\lVert H_\ell(x,a)_{T_x-1}\rVert_2$ and
$d_s=\lVert z_s-z_t\rVert_2^2$, the fixed weights are
\begin{equation}
w_s=\frac{(d_s+10^{-12})^{-1}}
{\sum_{r\in S}(d_r+10^{-12})^{-1}}.
\end{equation}
Each source condition has its own unedited-state weights. Both reuse routes from the
same source pair use the same numerical weights; comparisons across source conditions
use the same weighting rule, not necessarily the same values.
No edited target enters alignment or weighting.

\paragraph{Direct reuse.}
\label{app:direct-output-reuse}
With the same sources and weights, direct reuse predicts logits
\begin{equation*}
\widehat\ell_{t,e}=\ell(t,a)+\sum_s w_s[\ell(s,e)-\ell(s,a)].
\end{equation*}
Subtracting the unedited target logits gives its predicted output change.
Unlike hidden reuse, this route does not pass the source change through the
target's remaining layers.

\paragraph{Row-norm matching.}
For a nonzero donor row, magnitude matching uses
$\mathcal R(A,B)_j=\lVert B_j\rVert_2 A_j/\lVert A_j\rVert_2$.
A zero reference norm requires a zero matched row; a zero donor norm with positive
reference norm cannot be matched. This operation fixes row magnitudes, not content.

\paragraph{Distributional measures.}
For clean, edited and predicted logits $\ell_a,\ell_e,\ell_P$, set
$p_\bullet=\operatorname{softmax}(\ell_\bullet)$.
KL recovery is
\begin{equation}
R_{\mathrm{KL}}=1-
\frac{\mathrm{KL}(p_e\Vert p_P)}{\mathrm{KL}(p_e\Vert p_a)}.
\label{eq:kl-recovery}
\end{equation}
Values 1 and 0 denote exact reproduction and the clean target's error, respectively;
negative values denote larger error. Greedy agreement compares predicted and
edited-target top tokens, with the clean target as the zero-change baseline.
The cue-level aggregate in Table~\ref{tab:round11-readouts} uses the ratio of mean
block divergences, rather than a mean of per-target recovery ratios.

\paragraph{Complete-answer evaluation.}
\label{app:answer-reuse}
In the plain and plain+BOS formats, each question receives the cue
\texttt{\textbackslash nAnswer:\textbackslash n}, and candidates retain their released
wording and terminal newline. Chat renders a single user message containing the instruction
\emph{Answer with a short phrase only.} followed by a newline and the original question.
It uses the released template's generation prompt instead of this cue. The template's
default system content is retained, with Llama's default date fixed to 26 July 2024.
Each candidate ends with the family's end-of-turn token rather than a newline,
and that token is also scored.
For the rendered answer prompt $q^*$, a complete
candidate string $y$ has score
\begin{equation}
\ell(y\mid q^*)=\sum_{u=1}^{T_y}
\log p(y_u\mid q^*,y_{<u}).
\end{equation}
The sum uses full-vocabulary normalization with no length normalization.
The plain formats' newline-terminated sequences are prefix-free within each target's candidate set
on all three tokenizers. Exact duplicate strings enter the candidate set once.
For a published answer group $A$, the group score of Equation~\ref{eq:answer-response-error}
is $M_x(A)=\log\sum_{y\in A}\exp\ell(y\mid q^*)$ with $x=q^*$.
The measured contrast change $\delta c$ is the replacement-induced change in $M_x(A_T)-M_x(A_O)$.
Absolute error compares the predicted change with the directly measured edited
target's change, in nats; zero predicts no change.

At candidate position $u$, let $y_{<u}$ be the known preceding candidate tokens and
$s_a^*,s_e^*$ the rendered source prompts. The recorded response is
\begin{equation}
\Delta_{s,e}(y_{<u})=
H_\ell(s_e^*,y_{<u})-H_\ell(s_a^*,y_{<u}).
\label{eq:prefix-conditioned-response}
\end{equation}
Each prefix therefore has its own source capture, rather than reusing one bare-question
matrix for the whole answer. The clean target also receives that prefix.
The changed prompt suffix uses interpolation; shared candidate-prefix rows align tokenwise.
Source weights use the clean final prompt state and stay fixed across candidates.
For either reuse route, predicted logits at each prefix undergo full-vocabulary softmax,
giving $\widehat p_{t,e}(\cdot\mid y_{<u})$. Predicted complete-string and group scores are
\begin{equation}
\begin{aligned}
\widehat L_t(y)&=\sum_{u=1}^{T_y}
\log\widehat p_{t,e}(y_u\mid y_{<u}),\\
\widehat M_t(A)&=\log\sum_{y\in A}\exp\widehat L_t(y),\\
\widehat c_{t,e}&=\widehat M_t(A_T)-\widehat M_t(A_O).
\end{aligned}
\label{eq:predicted-answer-contrast}
\end{equation}
The predicted contrast change is $\widehat{\delta c}=\widehat c_{t,e}-c_{t,a}$, using the same
clean target contrast as the observed change in Equation~\ref{eq:answer-response-error}.

\FloatBarrier
\section{Study Materials}
\label{app:reuse-materials}

\subsection{Blocks and eligibility}
\label{app:crossed-confirmation}
Each answer-study block crosses two factual entity replacements with two queried
relations and six published ZsRE question templates: two sources and one target per
relation. Replacements preserve token length on all three tokenizers and change at
least two entity-overlapping tokens. Source and target contexts diverge before the
entity. Within each cohort, relation labels, entity identities and known aliases,
templates and complete inputs are disjoint across blocks; this does not make cohorts
independent of one another.

The introductory example and Table~\ref{tab:crossed-real-block} come from block 11 of
the earlier 40-block crossing, selected retrospectively for readable wording, not its
outcomes. They are not examples selected from the 35-block answer cohort.
The earlier next-token cohorts and their inference units are summarized in
Appendix~\ref{app:earlier-studies}.

\begin{table*}[!htbp]
\centering
\small
\caption{Six published question templates in one complete block. Substitute \(x\) with
each endpoint of SLIB to JDeveloper and Apache Derby to OpenLayers.
The entity slot is the only replacement in the displayed wording.}
\label{tab:crossed-real-block}
\begin{tabular}{@{}p{0.22\linewidth}l>{\raggedright\arraybackslash}p{0.59\linewidth}@{}}
\toprule
Relation label & Role & Published question \\
\midrule
developer & Source 1 & By whom was \(x\) developed? \\
developer & Source 2 & Which company developed \(x\)? \\
developer & Target & What is the name of \(x\)'s developer? \\
programming language & Source 1 & What is the programming language for \(x\)? \\
programming language & Source 2 & What programming language was used to write \(x\)? \\
programming language & Target & Which programming language is \(x\) written in? \\
\bottomrule
\end{tabular}
\end{table*}

\subsection{Answer cohorts and overlap}
\label{app:answer-confirmation}
The original answer cohort has 35 disjoint four-target blocks, 70 labels, 140 distinct
entities and 210 templates. It separates target- and other-relation new-answer strings
on all 140 targets, with no recorded dependency edges. Of the 70 relations and 210
templates, 68 and 202 recur from the crossing, and 67 and 197 recur from the
native-amplitude study. These are again new-entity rather than new-family materials.
Each target's candidates include published answers for both original and replacement
entities under both relations; only the replacement's groups $A_T$ and $A_O$ enter the
contrast, and the New-first and In-set summaries use the full candidate set
(Tables~\ref{tab:round10-bos} and~\ref{tab:round11-generation}). There are 532 target/candidate pairs per checkpoint,
at most six candidates per target. All sources, clean target states and predictions
are frozen for each checkpoint before edited-target answers are measured.

The two fresh cohorts contain 33 and 31 further four-target blocks. They reuse
64 of 66 and 62 of 62 relation labels and 192 of 198 and 185 of 186 templates from
earlier cohorts. Thus they assess new entities within largely familiar relation and
wording families, not new-family generalization. Selection excludes the earlier cohorts'
entities, aliases and complete inputs, retains equal-length, tokenizer-compatible
replacements, and uses the same block and answer-group construction. Only the input
rendering and scored terminator change across formats (Appendix~\ref{app:answer-reuse}).

\subsection{Artifacts, data and computation}
\label{app:artifacts}

\paragraph{Checkpoints and input formats.}
The fixed released checkpoints are Qwen2.5-0.5B-Instruct \citep{qwen2025qwen25},
hidden state 12 of 24 (dimension 896); Llama-3.2-1B-Instruct
\citep{meta2024llama32}, state 8 of 16 (dimension 2,048); and Gemma-2-2B-it
\citep{gemmateam2024gemma2}, state 13 of 26 (dimension 2,304), each at 50\% depth, the
interface at which changes are injected.
A scale replication uses the same-family Qwen2.5-7B-Instruct \citep{qwen2025qwen25},
Llama-3.1-8B-Instruct \citep{grattafiori2024llama3} and Gemma-2-9B-it
\citep{gemmateam2024gemma2}, at states 14 of 28,
16 of 32 and 21 of 42; their tokenizers are byte-identical to the small checkpoints', so the
answer inputs are unchanged.
Original studies use plain raw text without special tokens, fixed before outcomes:
no chat template or Llama/Gemma beginning-of-sequence (BOS) token, although both tokenizers
add one by default. The first token acts as an attention sink
\citep{xiao2024attentionsinks}; without it, one-token perturbations spread further
\citep{barbero2025firsttoken} and Gemma-2 loses its massive activations and degrades
\citep{owen2025massive}. Later protocols add each
checkpoint's chat
template and plain text with only the BOS token restored
(plain+BOS). The testbed, steering-completion and identical-content studies
use each checkpoint's default encoding, plain for Qwen, whose
tokenizer adds no BOS token, and plain+BOS for Llama and Gemma.

The checkpoints are used under their published terms: Apache 2.0 for
Qwen2.5-0.5B-Instruct, Qwen2.5-1.5B-Instruct and Qwen2.5-7B-Instruct, the Qwen Research
License, which permits non-commercial research use, for Qwen2.5-3B-Instruct, the Llama 3.2
and Llama 3.1 Community Licenses for Llama-3.2-1B-Instruct and Llama-3.1-8B-Instruct, and
the Gemma Terms of Use for both Gemma-2 checkpoints.
Materials come from public research releases of ZsRE, BEAR, T-REx (as distributed with
LPAQA), ParaRel and LPAQA, and from public Wikidata metadata. They state facts about
named public entities, including birth dates, family members and party membership;
no new personal data are collected. All prompts and answers are English. All model runs
use CPUs, float32 model weights and float64 scoring, with Python 3.12.9, PyTorch 2.13.0
\citep{paszke2019pytorch}, Transformers 5.14.1 \citep{wolf2020transformers} and NumPy 2.5.1
\citep{harris2020numpy}. The small, 1.5B and 3B checkpoints run with eight threads; the
larger checkpoints run as four 48-thread processes per 192-core node.
An earlier descriptive answer standard-patch extension on the original cohort, not analysed
here, and all 7B, 8B and 9B runs use the official PyTorch 2.13.0+cpu build and the
other measurements the macOS arm64 wheel of the same version, with the same scientific
settings and the $10^{-3}$ logit replay check: every continuation from an unpatched interface
must reproduce the clean logits within $10^{-3}$.
We plan to release the code, block and candidate identifiers, saved
predictions, run records and written protocols; text derived from ZsRE and LPAQA will
be released only as their terms permit.

BEAR is released under CC BY-SA 4.0, the ParaRel repository under MIT, and Wikidata
structured metadata under CC0.\footnote{Release terms: BEAR,
\url{https://github.com/lm-pub-quiz/BEAR}; ParaRel, \url{https://github.com/yanaiela/pararel};
Wikidata, \url{https://www.wikidata.org/wiki/Wikidata:Licensing}.}
We could not establish an explicit license for the exact ZsRE release or for
LPAQA's prompt files and bundled T-REx training subset from their release pages.\footnote{
ZsRE, \url{https://nlp.cs.washington.edu/zeroshot/}; LPAQA,
\url{https://github.com/jzbjyb/LPAQA}.}
Public availability is not treated as a redistribution license; reuse must retain
upstream attribution and resolve these terms before repackaging those materials.

The four earlier next-token cohorts and the original answer cohort total
2,076 target/checkpoint evaluations on the small checkpoints; the two fresh cohorts add 256 new targets, while auxiliary studies reuse
some targets. The answer standard-patch extension used 23.52 allocated core-hours plus 3.89
for an incomplete launch. The depth/scale, span-aligned-scan/chat/first-fresh-cohort,
BOS/neutral/second-cohort, testbed/depth, steering-completion and identical-content
studies used, respectively, about 3, 8, 6.5, 14.3, 4.3 and 1.8 reported CPU hours
for the smaller-checkpoint runs, and 448, 1,020, 1,178, 3,293, 889 and 404 allocated
core-hours for the larger checkpoints. The span-aligned-scan group also used 11.6
allocated core-hours in an incomplete launch; its smaller-checkpoint figure includes
the auxiliary descriptive 1.5B and 3B runs, whose outcomes are not analysed here.
CPU time and allocated core-hours are distinct
quantities; we did not record a project-wide total and do not infer one from target
counts. Saved-data checks run no models.

This work is intended for controlled analysis of response transfer, not deployment
of factual edits. Public named-entity facts can still be sensitive, outdated or biased;
identities are retained because entity replacement is the research variable, not
because the records are anonymous. We performed no dedicated offensive-content audit
and collected no new participant data. Response reuse could propagate false personal
claims or be misused to manipulate answers. The English-only selection and limited
checkpoints also preclude population-level fairness claims.

\subsection{Earlier next-token studies}
\label{app:earlier-studies}
Four studies preceding the answer evaluation measured next-token logit changes
immediately after the input prompt, without an answer cue or BOS token; the multi-token
study included 132 complete questions and 52 completion prompts among its 184 edits. They used the
three small checkpoints and public ZsRE, BEAR, T-REx, ParaRel and LPAQA materials
\citep{wiland2024bear,elsahar2018trex,elazar-etal-2021-measuring,jiang2020know}.
Table~\ref{tab:direction-studies} gives their sample sizes and inference units;
their 45 tests remain in the study-wide multiplicity accounting of
Table~\ref{tab:study-populations}.

\begin{table*}[!htbp]
\centering
\small
\caption{Earlier next-token studies in run order: role, main result, and targets and inference
units per checkpoint. Blocks contain four targets;
components group relations sharing templates.}
\label{tab:direction-studies}
\setlength{\tabcolsep}{3pt}
\begin{tabular}{@{}lllrl@{}}
\toprule
Study & Role & Main result & Targets & Units \\
\midrule
Single-token comparisons & Pre-spec. & Identity contrasts recur; derivative contrasts inconclusive & 64 & 10 components \\
Multi-token comparisons & Pre-spec. & Finite sources beat local derivatives & 184 & 45 components \\
Replacement/relation crossing & Pre-spec. & Same beats different replacement in 40 of 40 blocks & 160 & 40 blocks \\
Native amplitude & Pre-spec. & Route gain largely not replacement-specific & 144 & 36 blocks \\
\bottomrule
\end{tabular}
\end{table*}

Directional agreement is the cosine between predicted and measured logit-change
vectors after subtracting each vector's vocabulary mean. It is unchanged by positive
rescaling, unlike answer-contrast error. A target-local derivative is the Jacobian
of the input-to-interface computation at the original input applied to the full
embedding displacement of the replacement; a source-local derivative is computed
analogously at a source and then aligned and weighted. Neither is a path-integrated
derivative. Direction comparisons assign common row-norm profiles before exact
continuation through the target's remaining layers.

The single-token and multi-token studies each test four contrasts on three
checkpoints: finite source minus target-local derivative at the finite and derivative
profiles, same minus different replacement, and finite source minus source-local
derivative. At the finite profile the compared rows use finite-source magnitudes;
the derivative profile is globally rescaled to that total norm. Inference averages
relation medians equally within connected components of relations linked by shared
templates. The single-token sample has 64 BEAR replacements in ten relations and
ten components; the multi-token sample has 184 replacements in 46 relations and
45 components. All twelve multi-token comparisons pass their Holm family; only
the three identity contrasts pass in the single-token study, whose derivative
comparisons remain inconclusive at ten components.

The crossing uses 40 disjoint four-target blocks and a common same-replacement,
target-relation row-norm profile for every condition. Its nine tests compare same
with different replacements within each source relation and same/other with
different/target; all have 40 positive blocks. The native-amplitude study uses 36
further blocks and retains the measured source norms. Its twelve tests assess
the hidden-over-direct directional gain for the same replacement and the reduction
of the same-minus-different identity margin, on three checkpoints and two source
relations; all pass within their family. That margin reduction is specific to the
raw cosine scale: five of six mean contrasts change sign under Fisher-$z$,
squared-cosine or remaining-gap rescaling. These direction studies do not establish
complete-answer fidelity and do not replace the answer-error comparisons.

\subsection{Study sequence}
\label{app:study-sequence}
The independent single-token, multi-token, 40-block crossing and native-amplitude
studies preceded the 35-block answer study. An earlier 157-pair pilot is not evidence:
its pair-level tests assumed independence despite recurring subjects and relations.
Its identities, aliases and prompts are excluded from later selections. The pilot
fixed the 50\% interface; the later descriptive all-layer scan on the crossing cohort
did not retune it.

Each core study had an internal written protocol before its edited-target outcomes,
not an external registration. Later designs were informed by earlier findings:
the answer study chose other-relation sources after the crossing cohort's descriptive
answer comparison separated the routes most clearly. With target-relation sources,
the descriptive route-gap intervals included zero on Llama and Gemma.
After the plain-format answer
and scale studies came the span-aligned depth scan, chat format and the first fresh cohort.
That cohort's two pre-specified selectivity tests followed up the descriptive reductions
on Qwen2.5-7B and Llama-3.1-8B. BOS ablation, neutral sources and the second fresh cohort followed Gemma's known
plain-format failures. Subsequent protocols fixed the testbed, steering-completion
and identical-content comparisons before their own outcomes, but reused studied
materials, including already-observed Qwen blocks in some testbed arms.
Post-outcome diagnostics do not become pre-specified comparisons.

\FloatBarrier

\section{Prediction Fidelity, Calibration and the Answer Estimand}
\label{app:evidence-fidelity}
\label{app:prediction-quality}

The primary answer comparisons concern reproduction of a measured change, not factual
answer accuracy. Tables~\ref{tab:answer-errors} and~\ref{tab:answer-confirmation-tests}
report the absolute errors and the nine pre-specified answer comparisons against all
three baselines, including Gemma's positive mean gains with failed block sign tests.
Mean-effect intervals describe
magnitude; sign tests assess how often a block favors a comparison. The original answer
mean intervals use 20,000 outer-block percentile resamples with seed 20261007; later
tables state their aggregation and resampling scope.

\paragraph{Inference and test inventory.}
\label{app:statistical-details}
For a paired error comparison, positive effect means baseline error minus
evaluated-prediction error; two-sided construction differences state their
subtraction order in the table captions. Exact sign tests omit zero effects:
under the null, the number of positive signs among the nonzero blocks is
binomial with success probability one half. The one-sided test uses its upper
tail; the two-sided test doubles the smaller tail, capped at one. Empty or
incomplete planned comparisons receive $p=1$. Holm adjustment orders the raw
$p$ values within the stated family, multiplies each by the number of remaining
tests, and takes successive maxima capped at one. A named primary family does
not make the other families conditional on its success.

The original and larger-checkpoint answer families each contain nine tests
(three baselines on three checkpoints); chat has one nine-test family per scale.
The first fresh cohort adds those eighteen comparisons and two large-checkpoint
selectivity tests. The second adds six matched-route tests, eighteen answer
comparisons, two selectivity tests and twelve neutral-source tests
(other-relation versus neutral, and neutral versus zero).
BOS has one six-test Gemma family per cohort; Llama+BOS comparisons are descriptive.
The five testbed families, four steering families and two identical-content
control families each contain six tests, as defined in
Tables~\ref{tab:round11-families}, \ref{tab:b1-families} and~\ref{tab:controls-families}.
Together with the 45 earlier tests (Appendix~\ref{app:earlier-studies}), these form
the 223-test pooled sensitivity family, including two repeated comparisons.
Table~\ref{tab:round10-pooled} lists every conclusion altered by that pooling.
Power descriptions invert the exact binomial rejection rule at the first and
last Holm ranks for a true positive-sign probability; they do not use the
observed mean effect as a power estimate.
For the one-sided answer families with 31--35 nonzero blocks, the true
positive-sign probability needed for 80\% power ranges from about 0.71 to 0.80
across family sizes and the last to first Holm ranks. The larger testbed, steering
and identical-content control families state their sample sizes and rank-specific
thresholds in Tables~\ref{tab:round11-families}, \ref{tab:b1-families}
and~\ref{tab:controls-families}.

Mean intervals use percentile quantiles of 20,000 resamples of whole inference
units, preserving paired predictions. Pooled summaries resample blocks within
each of the 35/33/31-block cohorts. Label-resampled sensitivity intervals instead
sample the distinct relation labels with replacement and weight each block by
the product of its two labels' sampled counts, normalizing the weighted mean or
positive-block share. These post hoc route and content summaries use seed
20261013. The product-weight scheme can widen intervals even for independent
blocks: treating label counts as independent with mean and variance one gives
a heuristic variance multiplier near three, not a general correction factor.
It therefore measures sensitivity to this reweighting, not dependence-adjusted
family-wise significance. Median-block intervals invert the binomial sign test
to order statistics with coverage at least 0.95.

For route rescaling, each route gets a least-squares multiplicative factor
fitted on the original cohort, with no intercept; errors are then evaluated on
the two fresh cohorts only. This post hoc magnitude check is separate from the
observed-on-predicted calibration fits with an intercept in
Table~\ref{tab:answer-calibration}, and does not replace the original tests.

\begin{table}[!htbp]
\centering
\small
\caption{Mean answer-contrast error with other-relation sources in the plain format (nats;
lower is better) on 140 targets in 35 equally weighted blocks; zero predicts no change, and
same/different denotes replacement identity. Share removed is the fraction of the
zero-prediction error removed. $\dagger$: Holm-adjusted $p>0.05$ for same-replacement
hidden reuse against this baseline.}
\label{tab:answer-errors}
\setlength{\tabcolsep}{4pt}
\begin{tabular}{@{}lrrr@{}}
\toprule
Prediction & Qwen & Llama & Gemma \\
\midrule
\multicolumn{4}{@{}l}{(a) Small checkpoints} \\
Zero & 2.869 & 2.821 & $2.961^{\dagger}$ \\
Same, direct & 2.923 & 1.887 & 2.858 \\
Same, hidden & 1.572 & 1.427 & 2.298 \\
Different, hidden & 3.269 & 3.059 & $3.382^{\dagger}$ \\
Share removed, direct & $-2$\% & 33\% & 3\% \\
Share removed, hidden & 45\% & 49\% & 22\% \\
\addlinespace
\multicolumn{4}{@{}l}{(b) 7B, 8B and 9B checkpoints} \\
Zero & 6.974 & 3.905 & $3.385^{\dagger}$ \\
Same, direct & 8.923 & 4.536 & $3.793^{\dagger}$ \\
Same, hidden & 3.401 & 2.101 & 2.923 \\
Different, hidden & 6.978 & 3.829 & 3.880 \\
Share removed, direct & $-28$\% & $-16$\% & $-12$\% \\
Share removed, hidden & 51\% & 46\% & 14\% \\
\bottomrule
\end{tabular}
\end{table}
\begin{table*}[!htbp]
\centering
\small
\caption{Pre-specified answer-contrast error improvement over each baseline, in nats.
Positive means lower same-replacement/other-relation hidden error. Signs give
positive/negative/zero block counts. Intervals describe means pointwise; the corrected
sign tests assess prevalence of positive block effects.}
\label{tab:answer-confirmation-tests}
\begin{tabular}{@{}llrccr@{}}
\toprule
Model & Baseline & Mean & 95\% interval & Signs & $p_{\mathrm{Holm}}$ \\
\midrule
Qwen & Output & 1.351 & [0.882, 1.840] & 30/5/0 & $1.01\times10^{-4}$ \\
Qwen & Zero & 1.297 & [0.655, 2.073] & 28/7/0 & $1.52\times10^{-3}$ \\
Qwen & Different hidden & 1.696 & [1.107, 2.448] & 29/6/0 & $4.09\times10^{-4}$ \\
Llama & Output & 0.459 & [0.163, 0.767] & 27/8/0 & $4.70\times10^{-3}$ \\
Llama & Zero & 1.393 & [0.649, 2.295] & 26/9/0 & $1.20\times10^{-2}$ \\
Llama & Different hidden & 1.632 & [0.875, 2.570] & 30/5/0 & $1.01\times10^{-4}$ \\
Gemma & Output & 0.560 & [0.139, 0.977] & 25/10/0 & $2.50\times10^{-2}$ \\
Gemma & Zero & 0.663 & [0.077, 1.363] & 21/14/0 & $1.55\times10^{-1}$ \\
Gemma & Different hidden & 1.084 & [0.465, 1.822] & 23/12/0 & $8.95\times10^{-2}$ \\
\bottomrule
\end{tabular}
\end{table*}

\begin{figure*}[tbp]
\centering
\includegraphics[width=\textwidth]{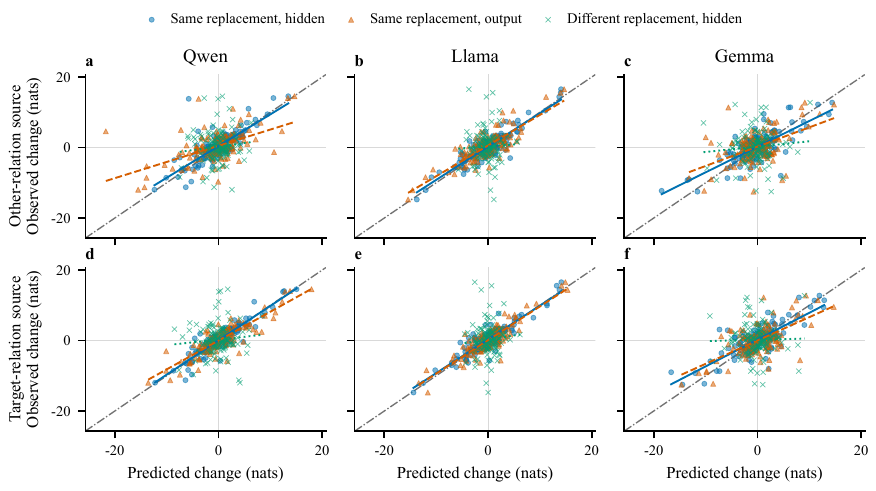}
\caption{Both source relations on shared axes, small checkpoints in the original
35-block cohort and plain format; all 140 targets per checkpoint and
three reuse conditions. Matching lines are calibration fits of observed on predicted
changes; the gray dash-dot line is exact reproduction. Target-relation sources bring
the output route closer to hidden reuse, without making the different-replacement
control track the target response.}
\label{fig:answer-prediction-both-relations}
\end{figure*}

\begin{figure*}[tbp]
\centering
\includegraphics[width=\textwidth]{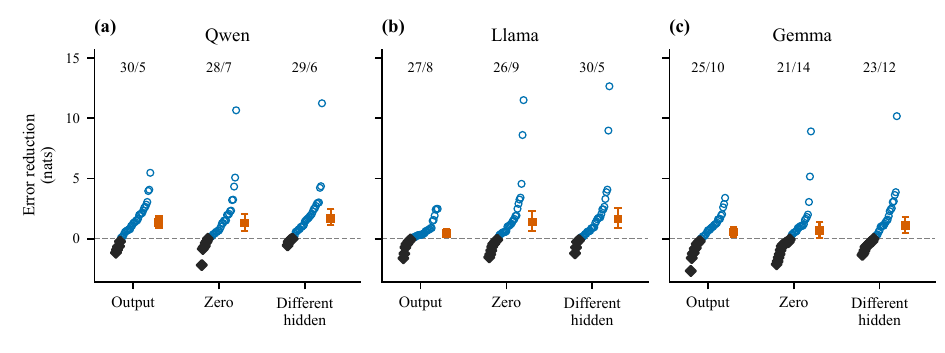}
\caption{All 35 block gains behind the nine original answer comparisons
(other-relation sources). Each point is one block's mean error reduction by
same-replacement hidden reuse relative to direct reuse, zero prediction or
different-replacement hidden reuse; open circles are positive and filled diamonds
negative. Squares and bars give block means and pointwise 95\% intervals; numbers give
positive/negative block counts.}
\label{fig:answer-block-gains}
\end{figure*}

\paragraph{Individual predictions and calibration.}
Figures~\ref{fig:answer-prediction-both-relations} and~\ref{fig:answer-block-gains}
show every target and block, respectively. Correlations measure covariation, whereas calibration regresses observed
change on predicted change with an intercept; no evaluated prediction is recalibrated.
A slope near one does not alone establish calibration, because intercept and mean bias
also matter (Table~\ref{tab:answer-calibration}). Table~\ref{tab:answer-paired-correlation} gives paired uncertainty for the
hidden-minus-direct correlation difference.

\begin{table*}[!htbp]
\centering
\small
\caption{Calibration of same-replacement predictions: observed change on predicted change.
Brackets give pointwise 95\% block-bootstrap intervals. Intercept and mean prediction
minus observed change (bias) are in nats. Other/target denotes the source relation.
Zero prediction has no calibration slope.}
\label{tab:answer-calibration}
\begin{tabular}{@{}lllrrr@{}}
\toprule
Source & Model & Route & Slope [95\%] & Intercept [95\%] & Bias [95\%] \\
\midrule
Other & Qwen & Hidden & 0.900 [0.622, 1.046] & 0.421 [0.047, 0.899] & $-$0.440 [$-$0.975, $-$0.048] \\
Other & Qwen & Output & 0.460 [0.130, 0.700] & 0.509 [$-$0.131, 1.160] & $-$0.812 [$-$1.905, 0.111] \\
Other & Llama & Hidden & 0.945 [0.787, 1.016] & 0.273 [$-$0.012, 0.553] & $-$0.271 [$-$0.556, 0.026] \\
Other & Llama & Output & 0.870 [0.667, 0.958] & 0.552 [0.078, 1.043] & $-$0.589 [$-$1.102, $-$0.092] \\
Other & Gemma & Hidden & 0.728 [0.560, 0.823] & 0.243 [$-$0.183, 0.701] & $-$0.253 [$-$0.781, 0.249] \\
Other & Gemma & Output & 0.548 [0.218, 0.779] & 0.253 [$-$0.285, 0.748] & $-$0.283 [$-$0.936, 0.365] \\
\addlinespace
Target & Qwen & Hidden & 0.976 [0.881, 1.045] & 0.076 [$-$0.131, 0.273] & $-$0.071 [$-$0.275, 0.135] \\
Target & Qwen & Output & 0.818 [0.698, 0.884] & $-$0.010 [$-$0.308, 0.305] & 0.068 [$-$0.275, 0.404] \\
Target & Llama & Hidden & 0.957 [0.820, 1.023] & 0.316 [0.047, 0.587] & $-$0.316 [$-$0.592, $-$0.046] \\
Target & Llama & Output & 0.953 [0.810, 1.025] & 0.130 [$-$0.233, 0.485] & $-$0.121 [$-$0.490, 0.244] \\
Target & Gemma & Hidden & 0.761 [0.573, 0.894] & 0.258 [$-$0.072, 0.591] & $-$0.271 [$-$0.608, 0.067] \\
Target & Gemma & Output & 0.662 [0.438, 0.818] & 0.010 [$-$0.388, 0.453] & 0.096 [$-$0.409, 0.587] \\
\addlinespace
\bottomrule
\end{tabular}
\end{table*}
\begin{table*}[!htbp]
\centering
\small
\caption{Paired Pearson-correlation differences for same-replacement hidden minus output
reuse. Intervals are pointwise descriptive, not new corrected tests. With target-relation
sources, Llama's interval includes zero.}
\label{tab:answer-paired-correlation}
\begin{tabular}{@{}llr@{}}
\toprule
Source & Model & $r_{\mathrm{hidden}}-r_{\mathrm{output}}$ [95\%] \\
\midrule
Other & Qwen & 0.249 [0.117, 0.415] \\
Other & Llama & 0.063 [0.017, 0.155] \\
Other & Gemma & 0.249 [0.139, 0.410] \\
Target & Qwen & 0.053 [0.013, 0.127] \\
Target & Llama & 0.023 [$-$0.008, 0.070] \\
Target & Gemma & 0.150 [0.055, 0.277] \\
\bottomrule
\end{tabular}
\end{table*}

\paragraph{Shared and relation-selective changes.}
\label{app:answer-structure}
The components $s$ and $h$ of Equation~\ref{eq:answer-paired-components} are algebraic
output summaries, not identified hidden-state mechanisms. Balanced pairing gives
\[
\operatorname{Var}(\delta c)=\operatorname{Var}(s)+\E[h^2].
\]
The selective share in Table~\ref{tab:core-component-variance} uses the first term;
the shared term is the second moment of $h$.
For component prediction errors
$e_s=\widehat s-s$ and $e_h=\widehat h-h$, the mean absolute error of a relation-swapped
target pair is
\[
\frac{|e_s+e_h|+|e_s-e_h|}{2}=\max(|e_s|,|e_h|).
\]
Thus good total response reproduction need not imply good relation-selective reproduction.
Table~\ref{tab:answer-components} reports both components' correlations and absolute
errors; selective-error reductions and their intervals appear in
Table~\ref{tab:selectivity-checks}, including nonpositive reductions.

\begin{table*}[!htbp]
\centering
\small
\caption{Selective share of observed contrast variance on the original 35 blocks,
plain format (\%). Balanced relation-swapped pairing uses
$\operatorname{Var}(\delta c)=\operatorname{Var}(s)+\E[h^2]$;
the share is $\operatorname{Var}(s)/\operatorname{Var}(\delta c)$, not
an attribution of the primary absolute-error gain. The complementary share is
shared between questions. Each checkpoint is evaluated on the same original cohort.}
\label{tab:core-component-variance}
\begin{tabular}{@{}lrrrrrr@{}}
\toprule
Checkpoint & Qwen 0.5B & Llama 1B & Gemma 2B & Qwen 7B & Llama 8B & Gemma 9B \\
\midrule
Selective share & 20.6 & 8.7 & 19.1 & 38.3 & 23.5 & 33.3 \\
\bottomrule
\end{tabular}
\end{table*}

\begin{table*}[!htbp]
\centering
\small
\caption{Paired contrast components: selective $s$ and shared $h$, 70 relation-swapped pairs
per checkpoint, retained in 35 four-target blocks. Errors are in nats; zero has undefined
correlation. Correlations use the fixed pair orientation described above.}
\label{tab:answer-components}
\begin{tabular}{@{}lllrrrr@{}}
\toprule
Source & Model & Prediction & $r_s$ & $r_h$ & MAE$_s$ & MAE$_h$ \\
\midrule
Other & Qwen & Same/hidden & 0.300 & 0.917 & 1.181 & 1.076 \\
Other & Qwen & Same/output & $-$0.499 & 0.885 & 2.538 & 1.508 \\
Other & Qwen & Different/hidden & 0.036 & 0.138 & 1.499 & 2.785 \\
Other & Qwen & Different/output & 0.017 & 0.039 & 2.057 & 3.154 \\
Other & Qwen & Zero & --- & --- & 1.293 & 2.571 \\
\addlinespace
Other & Llama & Same/hidden & 0.539 & 0.938 & 1.021 & 1.046 \\
Other & Llama & Same/output & $-$0.069 & 0.939 & 1.524 & 1.066 \\
Other & Llama & Different/hidden & 0.148 & 0.035 & 1.123 & 2.891 \\
Other & Llama & Different/output & 0.198 & $-$0.378 & 1.288 & 3.684 \\
Other & Llama & Zero & --- & --- & 0.916 & 2.720 \\
\addlinespace
Other & Gemma & Same/hidden & 0.383 & 0.828 & 1.562 & 1.673 \\
Other & Gemma & Same/output & $-$0.266 & 0.724 & 2.146 & 1.958 \\
Other & Gemma & Different/hidden & 0.224 & 0.004 & 1.582 & 2.943 \\
Other & Gemma & Different/output & $-$0.097 & $-$0.196 & 1.972 & 3.872 \\
Other & Gemma & Zero & --- & --- & 1.423 & 2.552 \\
\addlinespace
Target & Qwen & Same/hidden & 0.853 & 0.940 & 0.771 & 0.900 \\
Target & Qwen & Same/output & 0.686 & 0.919 & 1.197 & 1.194 \\
Target & Qwen & Different/hidden & 0.084 & 0.106 & 1.548 & 2.868 \\
Target & Qwen & Different/output & $-$0.032 & 0.031 & 1.942 & 3.027 \\
Target & Qwen & Zero & --- & --- & 1.293 & 2.571 \\
\addlinespace
Target & Llama & Same/hidden & 0.611 & 0.943 & 0.796 & 1.009 \\
Target & Llama & Same/output & 0.348 & 0.947 & 1.113 & 0.952 \\
Target & Llama & Different/hidden & 0.213 & 0.078 & 1.041 & 2.860 \\
Target & Llama & Different/output & $-$0.180 & $-$0.371 & 1.548 & 3.669 \\
Target & Llama & Zero & --- & --- & 0.916 & 2.720 \\
\addlinespace
Target & Gemma & Same/hidden & 0.413 & 0.865 & 1.371 & 1.443 \\
Target & Gemma & Same/output & 0.362 & 0.733 & 1.816 & 1.962 \\
Target & Gemma & Different/hidden & 0.108 & $-$0.057 & 1.620 & 3.052 \\
Target & Gemma & Different/output & 0.262 & $-$0.272 & 1.757 & 3.913 \\
Target & Gemma & Zero & --- & --- & 1.423 & 2.552 \\
\addlinespace
\bottomrule
\end{tabular}
\end{table*}

\paragraph{Candidate scope and sensitivity.}
\label{app:answer-candidate-mass}
The complete-string contrast uses full-vocabulary token normalization but only the literal
published strings in its answer groups. Their small absolute probability mass
(Table~\ref{tab:answer-candidate-mass}) limits an interpretation in terms of factual
recall or free generation. Table~\ref{tab:round12-robust} reports median-block summaries
and name-overlap exclusions, including the weakened plain-format Gemma results.
Table~\ref{tab:round12-routes} reports route rescaling and replacement-identity comparisons:
an advantage of the hidden route can also occur for a wrong replacement and is not by
itself evidence of replacement-specific transfer. These checks are post hoc descriptions,
not replacements for the original tests. Label-resampled intervals are sensitivity
summaries, not dependence-adjusted family-wise tests; interval widening can arise even
for independent blocks.

\begin{table*}[!htbp]
\centering
\small
\caption{Absolute complete-candidate mass: median $\log_{10}$ mass [25th, 75th percentiles]
over all 140 targets per small checkpoint in the original 35-block cohort and plain format.
Brackets describe the target distribution, not
confidence intervals. Other/target denotes source relation. All four nonzero predictions
per source relation are retained.}
\label{tab:answer-candidate-mass}
\begin{tabular}{@{}lrrr@{}}
\toprule
Endpoint or prediction & Qwen & Llama & Gemma \\
\midrule
Unedited target & $-$5.689 [$-$6.980, $-$4.429] & $-$4.607 [$-$5.380, $-$3.568] & $-$8.390 [$-$11.532, $-$6.821] \\
Replaced target & $-$5.744 [$-$7.164, $-$4.195] & $-$4.570 [$-$5.620, $-$3.421] & $-$8.397 [$-$11.195, $-$6.965] \\
Same/hidden, other & $-$5.625 [$-$7.149, $-$4.371] & $-$4.601 [$-$5.411, $-$3.350] & $-$8.387 [$-$11.252, $-$6.895] \\
Same/output, other & $-$5.866 [$-$7.399, $-$4.494] & $-$4.609 [$-$5.528, $-$3.354] & $-$8.692 [$-$11.158, $-$6.849] \\
Different/hidden, other & $-$5.954 [$-$7.217, $-$4.466] & $-$4.609 [$-$5.553, $-$3.515] & $-$8.542 [$-$11.021, $-$6.656] \\
Different/output, other & $-$6.195 [$-$7.879, $-$4.892] & $-$4.875 [$-$5.983, $-$3.864] & $-$8.873 [$-$11.430, $-$6.872] \\
Same/hidden, target & $-$5.780 [$-$7.323, $-$4.313] & $-$4.641 [$-$5.555, $-$3.306] & $-$8.241 [$-$10.944, $-$6.873] \\
Same/output, target & $-$5.909 [$-$7.402, $-$4.298] & $-$4.606 [$-$5.708, $-$3.340] & $-$8.796 [$-$11.397, $-$6.919] \\
Different/hidden, target & $-$5.747 [$-$7.169, $-$4.685] & $-$4.706 [$-$5.733, $-$3.462] & $-$8.645 [$-$11.193, $-$6.769] \\
Different/output, target & $-$6.256 [$-$7.826, $-$4.871] & $-$4.967 [$-$6.006, $-$3.790] & $-$8.956 [$-$11.764, $-$7.070] \\
\bottomrule
\end{tabular}
\end{table*}
\begin{table*}[!htbp]
\centering
\footnotesize
\caption{Robust summaries of the shares in Table~\ref{tab:answer-summary} (post hoc; no tests). For same-replacement hidden reuse with other-relation sources: the median over blocks of each block's share of its no-change error removed (Med.), the share after excluding targets whose answers share a word with an entity name (No ov.; 22, 8 and 14 targets in 8, 3 and 5 blocks of the three cohorts), and, over all 99 blocks, the share of Table~\ref{tab:answer-summary} with an interval resampling blocks within cohorts. Brackets on medians are distribution-free intervals from inverting the sign test (coverage at least 0.95). Chat rows cover the original blocks.}
\label{tab:round12-robust}
\setlength{\tabcolsep}{3pt}
\begin{tabular}{@{}llrrrrrrrrr@{}}
\toprule
 & & \multicolumn{2}{c}{Original} & \multicolumn{2}{c}{First fresh} & \multicolumn{2}{c}{Second fresh} & \multicolumn{3}{c}{All 99 blocks} \\
\cmidrule(lr){3-4}\cmidrule(lr){5-6}\cmidrule(lr){7-8}\cmidrule(lr){9-11}
Model & Input & Med. & No ov. & Med. & No ov. & Med. & No ov. & Share & Med. & No ov. \\
\midrule
Qwen 0.5B & plain & 35.1 & 35.3 & 32.3 & 40.0 & 30.2 & 35.7 & 41.9 [31.8, 50.5] & 32.3 [22.5, 47.4] & 37.1 \\
 & chat & 44.8 & 42.1 & -- & -- & -- & -- & -- & -- & -- \\
\addlinespace
Llama 1B & plain & 39.9 & 46.1 & 47.2 & 48.5 & 11.7 & 32.7 & 46.9 [38.0, 54.6] & 39.9 [29.8, 46.8] & 43.4 \\
 & +BOS & 46.2 & 52.4 & 64.8 & 63.5 & 52.7 & 56.6 & 59.7 [52.9, 65.6] & 55.7 [46.2, 62.6] & 57.7 \\
 & chat & 52.9 & 53.8 & -- & -- & -- & -- & -- & -- & -- \\
\addlinespace
Gemma 2B & plain & 20.5 & 7.8 & $-$6.1 & $-$9.1 & 19.3 & 15.3 & 14.6 [3.9, 24.4] & 11.6 [$-$7.4, 22.0] & 4.3 \\
 & +BOS & 63.1 & 65.4 & 60.1 & 63.2 & 60.6 & 56.1 & 63.9 [58.9, 68.2] & 61.3 [55.2, 64.5] & 61.8 \\
 & chat & 46.1 & 50.7 & -- & -- & -- & -- & -- & -- & -- \\
\addlinespace
Qwen 7B & plain & 55.6 & 54.2 & 45.5 & 38.9 & 51.1 & 50.9 & 47.5 [41.6, 52.9] & 51.3 [42.1, 55.6] & 47.8 \\
 & chat & 35.9 & 33.9 & -- & -- & -- & -- & -- & -- & -- \\
\addlinespace
Llama 8B & plain & 40.8 & 43.0 & 40.9 & 47.4 & 47.3 & 54.7 & 48.6 [42.7, 54.0] & 42.3 [36.1, 51.5] & 48.5 \\
 & +BOS & 38.5 & 40.6 & 46.1 & 48.1 & 55.7 & 55.2 & 48.5 [42.6, 53.9] & 45.2 [38.5, 54.9] & 48.3 \\
 & chat & 19.2 & 21.8 & -- & -- & -- & -- & -- & -- & -- \\
\addlinespace
Gemma 9B & plain & 12.1 & 5.3 & $-$26.6 & $-$18.8 & 3.7 & 7.5 & 6.4 [$-$4.5, 16.3] & 6.7 [$-$4.2, 14.7] & $-$1.5 \\
 & +BOS & 64.6 & 63.0 & 66.1 & 66.6 & 59.2 & 68.0 & 66.5 [62.2, 70.4] & 63.7 [56.6, 69.7] & 66.0 \\
 & chat & 51.6 & 54.6 & -- & -- & -- & -- & -- & -- & -- \\
\bottomrule
\end{tabular}
\end{table*}
\begin{table*}[!htbp]
\centering
\footnotesize
\caption{Route comparisons that allow for prediction magnitude and replacement identity (nats; post hoc). Unscaled and Rescaled: mean block error of direct minus hidden reuse (positive favors hidden reuse) on the two fresh cohorts (64 blocks), without and with each route's predictions multiplied by its least-squares factor fitted on the original cohort. Interaction: over all 99 blocks, the same-replacement route gain minus the different-replacement route gain; positive values mean a larger route gain for the same replacement. Brackets are 95\% intervals resampling blocks within cohorts.}
\label{tab:round12-routes}
\setlength{\tabcolsep}{5pt}
\begin{tabular}{@{}lllrrr@{}}
\toprule
Sources & Model & Input & Unscaled & Rescaled & Interaction \\
\midrule
Other relation & Qwen 0.5B & plain & 1.14 [0.80, 1.52] & 0.74 [0.47, 1.04] & 0.45 [0.07, 0.86] \\
 & Llama 1B & plain & 0.61 [0.41, 0.82] & 0.50 [0.32, 0.69] & $-$0.34 [$-$0.77, 0.05] \\
 & Gemma 2B & plain & 0.20 [$-$0.10, 0.51] & 0.13 [$-$0.07, 0.32] & $-$0.22 [$-$0.56, 0.10] \\
 & Qwen 7B & plain & 4.80 [3.82, 5.79] & 2.93 [2.26, 3.59] & 1.89 [0.69, 3.05] \\
 & Llama 8B & plain & 2.84 [2.23, 3.48] & 1.75 [1.37, 2.15] & 0.57 [$-$0.10, 1.22] \\
 & Gemma 9B & plain & 0.40 [0.09, 0.74] & 0.09 [$-$0.13, 0.32] & $-$0.21 [$-$0.67, 0.23] \\
 & Llama 1B & +BOS & 0.93 [0.73, 1.14] & 0.80 [0.61, 1.00] & $-$0.40 [$-$0.86, $-$0.01] \\
 & Gemma 2B & +BOS & 2.15 [1.65, 2.70] & 1.74 [1.30, 2.22] & 0.71 [0.08, 1.31] \\
 & Llama 8B & +BOS & 2.86 [2.24, 3.50] & 1.81 [1.40, 2.23] & 0.40 [$-$0.36, 1.11] \\
 & Gemma 9B & +BOS & 4.27 [3.41, 5.17] & 2.86 [2.28, 3.47] & 1.72 [0.78, 2.66] \\
\midrule
Target relation & Qwen 0.5B & plain & 0.23 [0.08, 0.39] & 0.24 [0.08, 0.41] & $-$0.42 [$-$0.78, $-$0.06] \\
 & Llama 1B & plain & 0.28 [0.12, 0.43] & 0.27 [0.12, 0.42] & $-$0.72 [$-$1.09, $-$0.39] \\
 & Gemma 2B & plain & 0.14 [$-$0.10, 0.37] & 0.11 [$-$0.07, 0.29] & $-$0.30 [$-$0.64, 0.03] \\
 & Qwen 7B & plain & 0.82 [0.33, 1.29] & 0.75 [0.31, 1.18] & $-$1.72 [$-$2.81, $-$0.69] \\
 & Llama 8B & plain & 0.44 [0.26, 0.63] & 0.40 [0.23, 0.57] & $-$1.45 [$-$2.05, $-$0.92] \\
 & Gemma 9B & plain & 0.36 [0.13, 0.60] & 0.22 [0.05, 0.39] & $-$0.24 [$-$0.70, 0.19] \\
 & Llama 1B & +BOS & 0.26 [0.14, 0.38] & 0.24 [0.12, 0.36] & $-$0.96 [$-$1.41, $-$0.57] \\
 & Gemma 2B & +BOS & 0.26 [0.11, 0.41] & 0.25 [0.10, 0.40] & $-$0.89 [$-$1.39, $-$0.46] \\
 & Llama 8B & +BOS & 0.36 [0.19, 0.55] & 0.38 [0.22, 0.55] & $-$1.66 [$-$2.26, $-$1.14] \\
 & Gemma 9B & +BOS & 0.55 [0.34, 0.76] & 0.61 [0.41, 0.82] & $-$1.72 [$-$2.50, $-$1.00] \\
\bottomrule
\end{tabular}
\end{table*}

\paragraph{Comparator and sensitivity details.}
Table~\ref{tab:core-extra-comparisons} separates the overlap exclusions, original-cohort
chat comparisons and 99-block construction differences. These scopes are not pooled
together. Table~\ref{tab:core-local-sensitivity} distinguishes the three-cohort source
comparison from the identical-content selectivity comparison and gives both mean
and positive-block-share intervals, including intervals crossing zero or one half.

\begin{table*}[!htbp]
\centering
\footnotesize
\caption{Additional comparisons for Sections~\ref{sec:answer-reproduction}
and~\ref{sec:testbed}, all at 1/2 depth. (a) Original plain-format overlap sensitivity:
positive/negative signs and Holm-adjusted $p$ after removing the eight name-overlap
blocks, leaving 27; the three baselines on three checkpoints retain one nine-test
Holm family per scale. This reduced analysis is post hoc. (b) Original 35-block chat cohort:
the share of no-change error removed by other-relation hidden or direct reuse (\%);
negative shares mean increased error. The last two columns are target-relation direct
error minus other-relation or target-relation hidden error (nats; pointwise 95\%
block-bootstrap interval; positive/negative blocks), so negative values favor
matched-source direct reuse; the last column gives the chat-format gaps of
Table~\ref{tab:answer-summary}. Neither route comparison was pre-specified.
(c) Descriptive paired error advantages (nats; pointwise 95\% intervals resampling
blocks within cohorts) over all 99 blocks in default encoding, with other-relation
sources and coefficient 1 for the entity-row vector. The last column gives the
native-scale vector's share of no-change error removed when broadcast from the
entity onward; negative shares mean larger mean error than zero prediction.
E denotes entity rows.}
\label{tab:core-extra-comparisons}
\textbf{(a) Name-overlap exclusions: signs; $p_{\mathrm{Holm}}$}\par\smallskip
\begin{tabular}{@{}lrrr@{}}
\toprule
Checkpoint & Versus direct & Versus zero & Versus different hidden \\
\midrule
Qwen 0.5B & 22/5; $6.1\times10^{-3}$ & 20/7; 0.038 & 22/5; $6.1\times10^{-3}$ \\
Llama 1B & 22/5; $6.1\times10^{-3}$ & 21/6; 0.015 & 24/3; $2.2\times10^{-4}$ \\
Gemma 2B & 19/8; 0.078 & 13/14; 0.701 & 15/12; 0.701 \\
Qwen 7B & 24/3; $1.7\times10^{-4}$ & 26/1; $1.9\times10^{-6}$ & 24/3; $1.7\times10^{-4}$ \\
Llama 8B & 25/2; $2.3\times10^{-5}$ & 22/5; $3.0\times10^{-3}$ & 24/3; $1.7\times10^{-4}$ \\
Gemma 9B & 16/11; 0.248 & 17/10; 0.248 & 22/5; $3.0\times10^{-3}$ \\
\bottomrule
\end{tabular}
\par\medskip
\textbf{(b) Chat-format comparators, original cohort}\par\smallskip
\begin{tabular}{@{}lrrrr@{}}
\toprule
Checkpoint & Hidden share & Direct share & \shortstack{Other hidden over\\target direct}
& \shortstack{Target hidden over\\target direct} \\
\midrule
Qwen 0.5B & 42.5 & $-$2.1 & $-$0.06 [$-$0.51, 0.39] 15/20 & 0.63 [0.25, 1.00] 22/13 \\
Llama 1B & 56.3 & 7.0 & $-$0.07 [$-$0.44, 0.28] 16/19 & 0.37 [0.03, 0.68] 26/9 \\
Gemma 2B & 46.4 & $-$19.9 & $-$1.08 [$-$1.60, $-$0.60] 11/24 & 0.34 [$-$0.11, 0.75] 23/12 \\
Qwen 7B & 35.2 & $-$67.6 & $-$0.88 [$-$2.12, 0.35] 16/19 & 1.20 [0.46, 1.94] 27/8 \\
Llama 8B & 20.0 & $-$63.4 & $-$3.08 [$-$4.29, $-$2.00] 8/27 & 0.21 [$-$0.24, 0.64] 20/15 \\
Gemma 9B & 52.4 & $-$77.5 & $-$1.60 [$-$2.43, $-$0.79] 9/26 & 0.20 [$-$0.50, 0.98] 21/14 \\
\bottomrule
\end{tabular}
\par\medskip
\textbf{(c) Coverage and alignment comparisons, 99 blocks}\par\smallskip
\begin{tabular}{@{}lrrr@{}}
\toprule
Checkpoint & \shortstack{Span over suffix\\interpolation}
& \shortstack{Span E over vector\\at each E row} & \shortstack{Native broadcast\\share (\%)} \\
\midrule
Qwen 0.5B & 0.20 [0.13, 0.27] & 0.45 [0.25, 0.67] & $-$239.8 \\
Llama 1B & 0.12 [0.06, 0.19] & 0.33 [0.13, 0.54] & $-$198.4 \\
Gemma 2B & 0.12 [0.05, 0.20] & 0.30 [0.08, 0.53] & $-$103.4 \\
Qwen 7B & 0.28 [0.07, 0.50] & 0.49 [0.12, 0.87] & $-$48.7 \\
Llama 8B & 0.18 [0.09, 0.28] & 0.30 [0.07, 0.55] & $-$82.0 \\
Gemma 9B & 0.14 [0.03, 0.26] & 0.37 [0.18, 0.56] & $-$38.2 \\
\bottomrule
\end{tabular}
\end{table*}

\begin{table*}[!htbp]
\centering
\footnotesize
\caption{Label-resampling sensitivity for two distinct 99-block comparisons,
with equal block weights in the point estimates (35 original, 33 first fresh,
31 second fresh). Brackets are saved 95\% intervals from 20,000 relation-label
resamples, seed 20261013; Fav.\ is the share of positive block effects, not a
probability of significance. (a) Plain format: neutral-source hidden error minus
other-relation hidden error. These pooled summaries are distinct from the
31-block tests in Table~\ref{tab:round10-neutral-route}.
(b) Default encoding: zero-prediction selective error minus error from the
same replacement's identical neutral entity-row content (D2a).
These intervals are post hoc sensitivity summaries, not adjusted sign tests.}
\label{tab:core-local-sensitivity}
\begin{tabular}{@{}lrr@{}}
\toprule
Checkpoint & Mean gain in nats [95\%] & Fav.\ [95\%] \\
\midrule
\multicolumn{3}{@{}l}{\textbf{(a) Other-relation versus neutral sources, plain format}} \\
Qwen 0.5B & 0.39 [0.10, 0.69] & 0.70 [0.52, 0.85] \\
Llama 1B & 0.45 [0.12, 0.84] & 0.66 [0.47, 0.82] \\
Gemma 2B & 0.06 [$-$0.32, 0.48] & 0.52 [0.33, 0.70] \\
Qwen 7B & 0.44 [$-$0.28, 1.18] & 0.57 [0.40, 0.72] \\
Llama 8B & 0.24 [$-$0.11, 0.57] & 0.63 [0.45, 0.78] \\
Gemma 9B & $-$0.09 [$-$0.57, 0.43] & 0.47 [0.30, 0.65] \\
\midrule
\multicolumn{3}{@{}l}{\textbf{(b) Identical-content selectivity, default encoding}} \\
Qwen 0.5B & 0.40 [0.13, 0.69] & 0.72 [0.54, 0.87] \\
Llama 1B & 0.16 [$-$0.03, 0.37] & 0.61 [0.44, 0.76] \\
Gemma 2B & 0.75 [0.38, 1.16] & 0.79 [0.62, 0.92] \\
Qwen 7B & 1.69 [0.81, 2.64] & 0.69 [0.50, 0.85] \\
Llama 8B & 0.74 [0.36, 1.20] & 0.70 [0.54, 0.84] \\
Gemma 9B & 1.29 [0.80, 1.90] & 0.84 [0.70, 0.94] \\
\bottomrule
\end{tabular}
\end{table*}

\FloatBarrier

\section{Replication, Encoding and Source-Relation Comparisons}
\label{app:evidence-replication}

These studies distinguish replication across checkpoint scales and input formats from
generalization to new relation families. The fresh cohorts contain new entities but
mostly familiar relations and templates. Tables~\ref{tab:depth-scale-confirmation},
\ref{tab:round9-families} and~\ref{tab:round10-families} report the larger-checkpoint,
chat-format and both fresh-cohort families, including failed comparisons.
Table~\ref{tab:round10-neutral-route} reports the second fresh cohort's failed
target-relation route tests and the neutral-source comparisons; failure does not show
equivalence. Appendix~\ref{app:artifacts} defines the input formats and
Appendix~\ref{app:study-sequence} gives their protocol order.

\paragraph{Matching the source relation.}
Table~\ref{tab:core-matched-routes} compares the two source conditions over the same
99 blocks and encoding.
Matching substantially narrows the mean route gap except for plain-format Gemma,
where the reduction is small. The source response and its read-out both change:
this is not an isolated causal effect of matching. The residual matched-route
advantages are descriptive, not confirmed by the second cohort's pre-specified tests;
the label-resampled intervals show the sensitivity behind this qualification.

\begin{table*}[!htbp]
\centering
\footnotesize
\caption{Source-relation matching at 1/2 depth with suffix interpolation on the same
99 blocks (35 original, 33 first fresh, 31 second fresh), equally weighted within
each encoding. Gap: direct minus hidden mean answer error (nats), using other- or
target-relation sources for the same replacement. Brackets: 95\% label-resampling
sensitivity intervals for the target mean gap and its share of favoring blocks
(Fav.), not family-wise tests. All six pre-specified matched-route tests on the
31-block second cohort failed (Table~\ref{tab:round10-neutral-route}).}
\label{tab:core-matched-routes}
\setlength{\tabcolsep}{4pt}
\begin{tabular}{@{}llrrr@{}}
\toprule
Checkpoint & Input & Other gap & Target gap [95\%] & Target Fav. [95\%] \\
\midrule
Qwen 0.5B & plain & 1.21 & 0.32 [0.10, 0.56] & 0.64 [0.48, 0.79] \\
Llama 1B & plain & 0.55 & 0.24 [0.03, 0.45] & 0.69 [0.51, 0.84] \\
Gemma 2B & plain & 0.33 & 0.29 [$-$0.09, 0.66] & 0.58 [0.41, 0.73] \\
Qwen 7B & plain & 5.05 & 0.90 [0.18, 1.56] & 0.70 [0.53, 0.84] \\
Llama 8B & plain & 2.70 & 0.40 [0.17, 0.65] & 0.73 [0.56, 0.87] \\
Gemma 9B & plain & 0.57 & 0.45 [0.09, 0.87] & 0.61 [0.44, 0.77] \\
Llama 1B & +BOS & 0.84 & 0.20 [0.05, 0.37] & 0.72 [0.57, 0.84] \\
Gemma 2B & +BOS & 1.97 & 0.26 [0.05, 0.45] & 0.68 [0.51, 0.83] \\
Llama 8B & +BOS & 2.67 & 0.36 [0.11, 0.64] & 0.73 [0.58, 0.86] \\
Gemma 9B & +BOS & 4.20 & 0.58 [0.26, 0.94] & 0.79 [0.64, 0.91] \\
\bottomrule
\end{tabular}
\end{table*}

\begin{table}[!htbp]
\centering
\small
\caption{Larger-checkpoint confirmation: same-replacement other-relation hidden reuse at
the 1/2 interface against three baselines on the 35 answer blocks. Gain is the mean
block error reduction (nats); signs count positive/negative blocks. Baselines are
direct reuse, zero prediction and different-replacement hidden reuse. One-sided exact
sign tests are Holm-adjusted within this nine-test family.}
\label{tab:depth-scale-confirmation}
\setlength{\tabcolsep}{3.5pt}
\begin{tabular}{@{}llrrr@{}}
\toprule
Checkpoint & Baseline & Signs & Gain & $p_{\mathrm{Holm}}$ \\
\midrule
Qwen 7B & Direct reuse & 32/3 & 5.523 & $1.7\times10^{-6}$ \\
Qwen 7B & Zero & 33/2 & 3.573 & $1.7\times10^{-7}$ \\
Qwen 7B & Different & 32/3 & 3.577 & $1.7\times10^{-6}$ \\
\addlinespace
Llama 8B & Direct reuse & 32/3 & 2.434 & $1.7\times10^{-6}$ \\
Llama 8B & Zero & 30/5 & 1.803 & $4.5\times10^{-5}$ \\
Llama 8B & Different & 32/3 & 1.728 & $1.7\times10^{-6}$ \\
\addlinespace
Gemma 9B & Direct reuse & 23/12 & 0.870 & 0.090 \\
Gemma 9B & Zero & 23/12 & 0.461 & 0.090 \\
Gemma 9B & Different & 29/6 & 0.957 & $1.8\times10^{-4}$ \\
\bottomrule
\end{tabular}
\end{table}
\begin{table*}[!htbp]
\centering
\footnotesize
\caption{The chat-format and first-fresh-cohort nine-comparison families: same-replacement other-relation hidden reuse at the 1/2 interface against direct reuse, zero prediction and different-replacement hidden reuse. Signs count positive/negative blocks (35 blocks in the chat format, 33 in the first fresh cohort); gains are mean block error reductions (nats); one-sided exact sign tests are Holm-adjusted within each family. $\dagger$: $p_{\mathrm{Holm}}>0.05$.}
\label{tab:round9-families}
\setlength{\tabcolsep}{2.5pt}
\begin{tabular}{@{}lrrrrrrrrr@{}}
\toprule
 & \multicolumn{3}{c}{Over direct reuse} & \multicolumn{3}{c}{Over zero} & \multicolumn{3}{c}{Over different} \\
\cmidrule(lr){2-4}\cmidrule(lr){5-7}\cmidrule(lr){8-10}
Checkpoint & Signs & Gain & $p_{\mathrm{Holm}}$ & Signs & Gain & $p_{\mathrm{Holm}}$ & Signs & Gain & $p_{\mathrm{Holm}}$ \\
\midrule
\multicolumn{10}{@{}l}{\emph{Chat format, small checkpoints}} \\
Qwen 0.5B & 29/6 & 1.92 & $1.2\times10^{-4}$ & 30/5 & 1.83 & $4.5\times10^{-5}$ & 27/8 & 2.03 & $9.4\times10^{-4}$ \\
Llama 1B & 30/5 & 2.14 & $4.5\times10^{-5}$ & 32/3 & 2.44 & $1.3\times10^{-6}$ & 34/1 & 3.43 & $9.4\times10^{-9}$ \\
Gemma 2B & 32/3 & 4.24 & $1.3\times10^{-6}$ & 33/2 & 2.97 & $1.5\times10^{-7}$ & 33/2 & 3.54 & $1.5\times10^{-7}$ \\
\addlinespace
\multicolumn{10}{@{}l}{\emph{Chat format, larger checkpoints}} \\
Qwen 7B & 32/3 & 8.99 & $1.7\times10^{-6}$ & 28/7 & 3.08 & 0.001 & 30/5 & 4.40 & $6.7\times10^{-5}$ \\
Llama 8B & 31/4 & 6.03 & $1.2\times10^{-5}$ & 25/10 & 1.45 & 0.008 & 26/9 & 2.04 & 0.006 \\
Gemma 9B & 34/1 & 12.13 & $9.4\times10^{-9}$ & 28/7 & 4.89 & 0.001 & 27/8 & 5.56 & 0.003 \\
\addlinespace
\multicolumn{10}{@{}l}{\emph{First fresh cohort, small checkpoints}} \\
Qwen 0.5B & 29/4 & 1.22 & $4.4\times10^{-5}$ & 25/8 & 1.05 & 0.009 & 29/4 & 1.63 & $4.4\times10^{-5}$ \\
Llama 1B & 28/5 & 0.58 & $1.7\times10^{-4}$ & 29/4 & 1.31 & $4.4\times10^{-5}$ & 30/3 & 1.17 & $6.3\times10^{-6}$ \\
Gemma 2B & 15/18 & $-$0.06 & 1.000$^{\dagger}$ & 15/18 & 0.05 & 1.000$^{\dagger}$ & 21/12 & 0.58 & 0.244$^{\dagger}$ \\
\addlinespace
\multicolumn{10}{@{}l}{\emph{First fresh cohort, larger checkpoints}} \\
Qwen 7B & 27/6 & 4.75 & $8.1\times10^{-4}$ & 26/7 & 3.12 & 0.003 & 28/5 & 3.47 & $2.0\times10^{-4}$ \\
Llama 8B & 29/4 & 3.00 & $4.4\times10^{-5}$ & 29/4 & 2.00 & $4.4\times10^{-5}$ & 31/2 & 2.18 & $5.9\times10^{-7}$ \\
Gemma 9B & 18/15 & 0.12 & 0.728$^{\dagger}$ & 15/18 & $-$0.17 & 0.757$^{\dagger}$ & 22/11 & 0.42 & 0.120$^{\dagger}$ \\
\bottomrule
\end{tabular}
\end{table*}
\begin{table*}[!htbp]
\centering
\footnotesize
\caption{Answer-comparison families of Appendix~\ref{app:evidence-replication}: same-replacement other-relation hidden reuse at the 1/2 interface against direct reuse, zero prediction and different-replacement hidden reuse. The plain+BOS families cover the two Gemma checkpoints, one six-test family per cohort; the second fresh cohort has one nine-test family per scale. Signs count positive/negative blocks; gains are mean block error reductions (nats); one-sided exact sign tests are Holm-adjusted within each family. $\dagger$: $p_{\mathrm{Holm}}>0.05$.}
\label{tab:round10-families}
\setlength{\tabcolsep}{2.5pt}
\begin{tabular}{@{}lrrrrrrrrr@{}}
\toprule
 & \multicolumn{3}{c}{Over direct reuse} & \multicolumn{3}{c}{Over zero} & \multicolumn{3}{c}{Over different} \\
\cmidrule(lr){2-4}\cmidrule(lr){5-7}\cmidrule(lr){8-10}
Checkpoint & Signs & Gain & $p_{\mathrm{Holm}}$ & Signs & Gain & $p_{\mathrm{Holm}}$ & Signs & Gain & $p_{\mathrm{Holm}}$ \\
\midrule
\multicolumn{10}{@{}l}{\emph{Plain+BOS, original blocks (35 blocks)}} \\
Gemma 2B & 32/3 & 1.63 & $4.2\times10^{-7}$ & 35/0 & 2.92 & $1.7\times10^{-10}$ & 32/3 & 2.93 & $4.2\times10^{-7}$ \\
Gemma 9B & 34/1 & 4.08 & $4.2\times10^{-9}$ & 34/1 & 3.00 & $4.2\times10^{-9}$ & 35/0 & 3.20 & $1.7\times10^{-10}$ \\
\addlinespace
\multicolumn{10}{@{}l}{\emph{Plain+BOS, first fresh cohort (33 blocks)}} \\
Gemma 2B & 30/3 & 2.21 & $1.4\times10^{-6}$ & 31/2 & 2.35 & $2.0\times10^{-7}$ & 29/4 & 2.16 & $5.5\times10^{-6}$ \\
Gemma 9B & 32/1 & 4.26 & $2.0\times10^{-8}$ & 32/1 & 3.28 & $2.0\times10^{-8}$ & 33/0 & 3.23 & $7.0\times10^{-10}$ \\
\addlinespace
\multicolumn{10}{@{}l}{\emph{Plain+BOS, second fresh cohort (31 blocks)}} \\
Gemma 2B & 29/2 & 2.10 & $6.9\times10^{-7}$ & 28/3 & 1.99 & $4.6\times10^{-6}$ & 30/1 & 2.25 & $8.9\times10^{-8}$ \\
Gemma 9B & 30/1 & 4.27 & $8.9\times10^{-8}$ & 28/3 & 3.49 & $4.6\times10^{-6}$ & 30/1 & 3.16 & $8.9\times10^{-8}$ \\
\addlinespace
\multicolumn{10}{@{}l}{\emph{Plain format, second fresh cohort (31 blocks), small checkpoints}} \\
Qwen 0.5B & 24/7 & 1.04 & 0.012 & 22/9 & 0.94 & 0.074$^{\dagger}$ & 25/6 & 1.29 & 0.004 \\
Llama 1B & 22/9 & 0.64 & 0.074$^{\dagger}$ & 21/10 & 0.88 & 0.074$^{\dagger}$ & 28/3 & 1.57 & $2.1\times10^{-5}$ \\
Gemma 2B & 22/9 & 0.49 & 0.074$^{\dagger}$ & 20/11 & 0.44 & 0.075$^{\dagger}$ & 23/8 & 1.00 & 0.032 \\
\addlinespace
\multicolumn{10}{@{}l}{\emph{Plain format, second fresh cohort (31 blocks), larger checkpoints}} \\
Qwen 7B & 27/4 & 4.85 & $1.0\times10^{-4}$ & 27/4 & 3.65 & $1.0\times10^{-4}$ & 28/3 & 3.41 & $2.1\times10^{-5}$ \\
Llama 8B & 28/3 & 2.67 & $2.1\times10^{-5}$ & 28/3 & 2.13 & $2.1\times10^{-5}$ & 27/4 & 2.28 & $1.0\times10^{-4}$ \\
Gemma 9B & 23/8 & 0.70 & 0.016 & 17/14 & 0.28 & 0.720$^{\dagger}$ & 17/14 & 0.60 & 0.720$^{\dagger}$ \\
\bottomrule
\end{tabular}
\end{table*}
\begin{table*}[!htbp]
\centering
\footnotesize
\caption{Neutral-source and read-out-matched route families on the second fresh cohort (31 blocks, plain format). Other over neutral compares same-replacement hidden reuse from other-relation sources with hidden reuse from the two relation-neutral sources; neutral over zero compares the neutral sources with zero prediction (one family of six per scale). The matched route compares hidden with direct reuse when both use target-relation sources (one family of six). Signs, gains (nats) and Holm adjustment as in Table~\ref{tab:round10-families}. $\dagger$: $p_{\mathrm{Holm}}>0.05$.}
\label{tab:round10-neutral-route}
\setlength{\tabcolsep}{2.5pt}
\begin{tabular}{@{}lrrrrrrrrr@{}}
\toprule
 & \multicolumn{3}{c}{Other over neutral} & \multicolumn{3}{c}{Neutral over zero} & \multicolumn{3}{c}{Matched route} \\
\cmidrule(lr){2-4}\cmidrule(lr){5-7}\cmidrule(lr){8-10}
Checkpoint & Signs & Gain & $p_{\mathrm{Holm}}$ & Signs & Gain & $p_{\mathrm{Holm}}$ & Signs & Gain & $p_{\mathrm{Holm}}$ \\
\midrule
Qwen 0.5B & 21/10 & 0.40 & 0.142$^{\dagger}$ & 19/12 & 0.54 & 0.422$^{\dagger}$ & 21/10 & 0.38 & 0.212$^{\dagger}$ \\
Llama 1B & 22/9 & 0.45 & 0.088$^{\dagger}$ & 19/12 & 0.43 & 0.422$^{\dagger}$ & 19/12 & 0.18 & 0.422$^{\dagger}$ \\
Gemma 2B & 22/9 & 0.47 & 0.088$^{\dagger}$ & 19/12 & $-$0.03 & 0.422$^{\dagger}$ & 14/17 & $-$0.07 & 0.763$^{\dagger}$ \\
\addlinespace
Qwen 7B & 18/13 & 0.35 & 0.710$^{\dagger}$ & 26/5 & 3.30 & $5.8\times10^{-4}$ & 21/10 & 0.41 & 0.212$^{\dagger}$ \\
Llama 8B & 19/12 & 0.22 & 0.562$^{\dagger}$ & 25/6 & 1.91 & 0.002 & 21/10 & 0.34 & 0.212$^{\dagger}$ \\
Gemma 9B & 14/17 & 0.07 & 1.000$^{\dagger}$ & 16/15 & 0.22 & 1.000$^{\dagger}$ & 18/13 & 0.51 & 0.473$^{\dagger}$ \\
\bottomrule
\end{tabular}
\end{table*}

\paragraph{Neutral-source shares by cohort.}
\label{app:neutral-cohort-details}
In the plain format, the other-relation/neutral-source hidden shares of no-change error removed are 51.2/39.6, 42.4/40.9, 48.7/44.1\% for Qwen2.5-7B and 46.2/44.2, 49.3/38.7, 50.4/45.1\% for Llama-3.1-8B, in original/first-fresh/second-fresh order (35/33/31 blocks). These are cohort-specific ratios, distinct from the pooled nats in Table~\ref{tab:core-local-sensitivity}.

Restoring only the BOS token is distinct from changing to a chat template.
Table~\ref{tab:round10-bos} reports the format-specific results rather than pooling away
Gemma's plain-format failures. Table~\ref{tab:selectivity-checks} shows why the selectivity
sign tests alone are insufficient: a constant push toward each question's own answers
would also pass the four specified fresh-cohort tests. The magnitudes and
different-replacement comparison carry additional information. The study-wide Holm
sensitivity in Table~\ref{tab:round10-pooled} lists every conclusion that changes when
all 223 tests, which include the two repeated comparisons, are pooled.

\begin{table*}[!htbp]
\centering
\footnotesize
\caption{Plain format against plain+BOS on the three cohorts (descriptive except the Gemma signs, which are the tests of Table~\ref{tab:round10-families}). Each pair gives plain / plain+BOS values of the mean zero-prediction error (nats), the median log$_{10}$ candidate mass, the share of the zero error removed by same-replacement other-relation hidden reuse (\%), the reduction in selective absolute error relative to zero (\%), and the number of targets whose highest-scoring edited candidate is a published target-relation answer of the replacement (of 140, 132 and 124). The last column gives plain+BOS block signs over direct reuse, zero prediction and different-replacement hidden reuse; Llama's are not tested.}
\label{tab:round10-bos}
\setlength{\tabcolsep}{3pt}
\begin{tabular}{@{}llrrrrrr@{}}
\toprule
Checkpoint & Cohort & Zero error & Mass & Share & Selective & New first & Signs, plain+BOS \\
\midrule
Llama 1B & Original & 2.82 / 2.81 & $-$4.61 / $-$4.61 & 49.4 / 56.5 & $-$11.4 / 8.3 & 82 / 76 & 27/8, 31/4, 31/4 \\
 & First fresh & 2.63 / 2.62 & $-$4.86 / $-$4.76 & 50.0 / 66.0 & 21.2 / 41.4 & 65 / 66 & 29/4, 30/3, 31/2 \\
 & Second fresh & 2.22 / 2.25 & $-$4.39 / $-$4.44 & 39.6 / 56.5 & 0.4 / 18.4 & 52 / 60 & 28/3, 28/3, 29/2 \\
\addlinespace
Gemma 2B & Original & 2.96 / 4.31 & $-$8.39 / $-$4.53 & 22.4 / 67.8 & $-$9.8 / 35.9 & 42 / 91 & 32/3, 35/0, 32/3 \\
 & First fresh & 2.53 / 3.63 & $-$8.97 / $-$4.31 & 2.1 / 64.6 & $-$32.9 / 52.0 & 36 / 83 & 30/3, 31/2, 29/4 \\
 & Second fresh & 2.49 / 3.45 & $-$8.25 / $-$4.16 & 17.8 / 57.6 & $-$8.8 / 40.0 & 38 / 74 & 29/2, 28/3, 30/1 \\
\addlinespace
Llama 8B & Original & 3.90 / 4.21 & $-$3.09 / $-$3.37 & 46.2 / 40.8 & 19.7 / 2.6 & 101 / 100 & 29/6, 30/5, 30/5 \\
 & First fresh & 4.07 / 4.40 & $-$2.98 / $-$3.16 & 49.3 / 50.8 & 35.0 / 26.4 & 97 / 101 & 28/5, 29/4, 29/4 \\
 & Second fresh & 4.23 / 4.90 & $-$2.75 / $-$3.00 & 50.4 / 53.8 & 34.1 / 35.7 & 89 / 95 & 29/2, 25/6, 26/5 \\
\addlinespace
Gemma 9B & Original & 3.38 / 4.62 & $-$8.05 / $-$4.66 & 13.6 / 65.0 & $-$7.3 / 48.7 & 49 / 99 & 34/1, 34/1, 35/0 \\
 & First fresh & 2.71 / 4.76 & $-$8.36 / $-$4.45 & $-$6.3 / 68.8 & $-$21.9 / 57.2 & 41 / 97 & 32/1, 32/1, 33/0 \\
 & Second fresh & 3.00 / 5.30 & $-$7.87 / $-$4.29 & 9.5 / 65.8 & $-$23.5 / 54.3 & 38 / 85 & 30/1, 28/3, 30/1 \\
\bottomrule
\end{tabular}
\end{table*}
\begin{table*}[!htbp]
\centering
\footnotesize
\caption{Selectivity checks of same-replacement other-relation hidden reuse (post hoc; no tests). For each listed cohort, encoding and checkpoint: the share of relation-swapped pairs whose observed selective component $s$ is positive (\%); blocks favoring / opposing a vanishingly small constant $\widehat s>0$ under the selectivity statistic $\lvert s\rvert-\lvert\widehat s-s\rvert$ (Push; blocks whose two pairs differ in sign tie); the block signs of that statistic for hidden reuse (Hidden); the reduction in selective absolute error relative to zero prediction (\%) of same-replacement (Same) and different-replacement (Diff.) hidden reuse, with 95\% block-bootstrap intervals; and the mean block advantage in selective absolute error of the same over the different replacement (nats), with its interval and favoring / opposing blocks. In the pre-specified two-test selectivity families (plain format, Qwen 7B and Llama 8B), the constant push would give Holm-adjusted $p$ of 0.011 and 0.002 on the first fresh cohort and 0.029 and $<$0.001 on the second; hidden reuse gives 0.001 and 0.018, and 0.075 and $<$0.001. The final panel reports the original 35-block chat selective-error reductions with pointwise 95\% block-bootstrap intervals; these are descriptive, not multiplicity-adjusted tests.}
\label{tab:selectivity-checks}
\setlength{\tabcolsep}{4pt}
\begin{tabular}{@{}lrrrrrr@{}}
\toprule
Checkpoint & $s>0$ & Push & Hidden & Same & Diff. & Same over different \\
\midrule
\multicolumn{7}{@{}l}{\emph{Original blocks, plain format}} \\
Qwen 0.5B & 52.9 & 11/9 & 19/16 & 8.7 [$-$13.5, 28.9] & $-$15.9 [$-$32.9, 1.7] & 0.32 [0.01, 0.63] 21/14 \\
Llama 1B & 58.6 & 15/9 & 14/21 & $-$11.4 [$-$42.7, 9.9] & $-$22.6 [$-$46.9, $-$3.2] & 0.10 [$-$0.10, 0.32] 17/18 \\
Gemma 2B & 50.0 & 8/8 & 14/21 & $-$9.8 [$-$33.9, 10.3] & $-$11.2 [$-$28.1, 3.8] & 0.02 [$-$0.31, 0.36] 19/16 \\
Qwen 7B & 62.9 & 15/6 & 27/8 & 50.2 [36.8, 60.8] & 3.5 [$-$16.8, 20.2] & 2.16 [1.28, 3.06] 27/8 \\
Llama 8B & 74.3 & 22/5 & 23/12 & 19.7 [1.5, 34.2] & $-$10.5 [$-$28.3, 3.0] & 0.65 [0.31, 1.00] 28/7 \\
Gemma 9B & 48.6 & 9/10 & 16/19 & $-$7.3 [$-$25.5, 11.1] & $-$7.4 [$-$28.6, 11.5] & 0.00 [$-$0.59, 0.56] 20/15 \\
\addlinespace
\multicolumn{7}{@{}l}{\emph{First fresh cohort, plain format}} \\
Qwen 0.5B & 62.1 & 13/5 & 23/10 & 17.2 [$-$1.6, 32.9] & $-$12.6 [$-$33.5, 5.8] & 0.44 [0.05, 0.85] 21/12 \\
Llama 1B & 53.0 & 11/9 & 20/13 & 21.2 [$-$0.4, 38.3] & 0.1 [$-$19.0, 14.5] & 0.26 [$-$0.03, 0.55] 18/15 \\
Gemma 2B & 54.5 & 8/5 & 7/26 & $-$32.9 [$-$58.3, $-$7.3] & $-$10.0 [$-$35.0, 10.2] & $-$0.32 [$-$0.80, 0.09] 17/16 \\
Qwen 7B & 65.2 & 13/3 & 26/7 & 35.1 [17.7, 49.0] & $-$9.4 [$-$30.7, 7.2] & 1.99 [0.97, 3.05] 24/9 \\
Llama 8B & 74.2 & 20/4 & 23/10 & 35.0 [21.4, 46.5] & $-$1.8 [$-$21.7, 15.2] & 0.93 [0.50, 1.37] 27/6 \\
Gemma 9B & 43.9 & 5/9 & 15/18 & $-$21.9 [$-$47.1, $-$1.5] & $-$8.8 [$-$26.6, 7.6] & $-$0.20 [$-$0.64, 0.23] 15/18 \\
\addlinespace
\multicolumn{7}{@{}l}{\emph{Second fresh cohort, plain format}} \\
Qwen 0.5B & 66.1 & 15/5 & 23/8 & 21.8 [$-$0.4, 40.7] & $-$7.3 [$-$27.6, 8.8] & 0.39 [0.05, 0.74] 22/9 \\
Llama 1B & 51.6 & 8/7 & 18/13 & 0.4 [$-$28.4, 21.0] & $-$34.7 [$-$68.1, $-$9.5] & 0.29 [0.03, 0.57] 19/12 \\
Gemma 2B & 41.9 & 5/10 & 14/17 & $-$8.8 [$-$30.6, 8.9] & $-$18.6 [$-$44.5, 2.2] & 0.13 [$-$0.20, 0.45] 19/12 \\
Qwen 7B & 62.9 & 11/3 & 20/11 & 36.5 [14.0, 52.8] & 3.6 [$-$18.2, 20.4] & 1.43 [0.43, 2.43] 20/11 \\
Llama 8B & 80.6 & 20/1 & 26/5 & 34.1 [20.7, 45.2] & 4.6 [$-$12.3, 21.0] & 0.69 [0.21, 1.16] 20/11 \\
Gemma 9B & 46.8 & 4/6 & 9/22 & $-$23.5 [$-$52.8, 0.3] & $-$21.8 [$-$48.3, 1.6] & $-$0.03 [$-$0.55, 0.50] 15/16 \\
\addlinespace
\multicolumn{7}{@{}l}{\emph{Original blocks, plain+BOS}} \\
Llama 1B & 61.4 & 12/4 & 17/18 & 8.3 [$-$20.4, 31.2] & $-$19.9 [$-$45.1, 0.6] & 0.26 [0.03, 0.51] 21/14 \\
Gemma 2B & 55.7 & 10/6 & 25/10 & 35.9 [20.9, 48.2] & $-$9.4 [$-$27.9, 6.0] & 0.76 [0.44, 1.09] 25/10 \\
Llama 8B & 81.4 & 25/3 & 16/19 & 2.6 [$-$20.4, 20.0] & $-$9.6 [$-$31.9, 8.4] & 0.27 [$-$0.17, 0.72] 19/16 \\
Gemma 9B & 72.9 & 20/4 & 27/8 & 48.7 [32.4, 61.4] & $-$3.5 [$-$17.1, 8.3] & 1.21 [0.76, 1.66] 28/7 \\
\addlinespace
\multicolumn{7}{@{}l}{\emph{First fresh cohort, plain+BOS}} \\
Llama 1B & 60.6 & 11/4 & 24/9 & 41.4 [20.7, 55.3] & $-$9.7 [$-$28.7, 5.4] & 0.48 [0.26, 0.71] 26/7 \\
Gemma 2B & 62.1 & 13/5 & 28/5 & 52.0 [36.5, 63.3] & $-$0.3 [$-$14.6, 12.3] & 0.97 [0.53, 1.47] 23/10 \\
Llama 8B & 71.2 & 17/3 & 24/9 & 26.4 [7.0, 40.8] & 3.0 [$-$20.9, 21.2] & 0.60 [0.13, 1.13] 22/11 \\
Gemma 9B & 74.2 & 17/1 & 29/4 & 57.2 [45.9, 65.7] & 0.4 [$-$15.9, 14.2] & 1.59 [1.01, 2.28] 30/3 \\
\addlinespace
\multicolumn{7}{@{}l}{\emph{Second fresh cohort, plain+BOS}} \\
Llama 1B & 58.1 & 11/6 & 18/13 & 18.4 [$-$4.0, 37.3] & $-$34.7 [$-$58.0, $-$17.4] & 0.43 [0.25, 0.61] 24/7 \\
Gemma 2B & 59.7 & 11/5 & 22/9 & 40.0 [20.2, 55.8] & $-$15.5 [$-$27.6, $-$4.5] & 0.96 [0.61, 1.32] 26/5 \\
Llama 8B & 75.8 & 20/4 & 23/8 & 35.7 [19.7, 48.5] & 1.9 [$-$12.3, 14.2] & 0.94 [0.44, 1.42] 26/5 \\
Gemma 9B & 75.8 & 18/2 & 26/5 & 54.3 [42.7, 64.5] & $-$1.6 [$-$16.8, 11.7] & 1.49 [1.03, 1.96] 29/2 \\
\bottomrule
\end{tabular}
\par\medskip
\begin{tabular}{@{}lr@{}}
\toprule
Checkpoint & Selective error reduction (\%) [95\%] \\
\midrule
\multicolumn{2}{@{}l}{\emph{Original 35 blocks, chat format}} \\
Qwen 0.5B & 21.8 [1.2, 38.1] \\
Llama 1B & 40.7 [26.5, 51.7] \\
Gemma 2B & 38.3 [22.0, 51.4] \\
Qwen 7B & 37.2 [20.4, 51.0] \\
Llama 8B & 6.7 [$-$10.4, 19.9] \\
Gemma 9B & 52.1 [39.9, 61.7] \\
\bottomrule
\end{tabular}
\end{table*}
\begin{table*}[!htbp]
\centering
\footnotesize
\caption{All 26 conclusions that change under the final 223-test pooled Holm
correction: original family-wise and pooled adjusted $p$ values. Every other test
keeps its conclusion; Table~\ref{tab:study-populations} gives the study inventory.}
\label{tab:round10-pooled}
\setlength{\tabcolsep}{5pt}
\begin{tabular}{@{}lrr@{}}
\toprule
Test & Family $p_{\mathrm{Holm}}$ & Pooled (223) \\
\midrule
Single-token: Qwen same over different replacement & 0.012 & 0.081 \\
Single-token: Llama same over different replacement & 0.012 & 0.081 \\
Single-token: Gemma same over different replacement & 0.012 & 0.081 \\
Answer study, small, plain: Llama 1B over direct reuse & 0.005 & 0.081 \\
Answer study, small, plain: Llama 1B over zero & 0.012 & 0.225 \\
Answer study, small, plain: Gemma 2B over direct reuse & 0.025 & 0.592 \\
Chat format, small: Qwen 0.5B over different & $9.4\times10^{-4}$ & 0.081 \\
Chat format, larger: Llama 8B over zero & 0.008 & 0.592 \\
Chat format, larger: Llama 8B over different & 0.006 & 0.225 \\
Chat format, larger: Gemma 9B over different & 0.003 & 0.081 \\
First fresh cohort, small: Qwen 0.5B over zero & 0.009 & 0.175 \\
First fresh cohort, larger: Qwen 7B over zero & 0.003 & 0.059 \\
First fresh cohort, selectivity: Qwen 7B & 0.001 & 0.059 \\
First fresh cohort, selectivity: Llama 8B & 0.018 & 0.947 \\
Second fresh cohort, larger: Gemma 9B over direct reuse & 0.016 & 0.390 \\
Second fresh cohort, small: Gemma 2B over different & 0.032 & 0.390 \\
Second fresh cohort, small: Qwen 0.5B over direct reuse & 0.012 & 0.131 \\
Testbed, span over tuned vector: Llama 8B & 0.034 & 0.944 \\
Testbed, span over tuned vector: Qwen 7B & 0.034 & 0.944 \\
Testbed, identical-content selectivity: Llama 1B & 0.022 & 1.000 \\
Steering, tuned later-row vector over zero: Llama 1B & 0.007 & 0.144 \\
Steering, tuned later-row vector over zero: Llama 8B & 0.004 & 0.068 \\
Steering, span over best vector: Qwen 7B & 0.003 & 0.125 \\
Steering, norm-matched later rows over zero: Llama 1B & 0.001 & 0.053 \\
Identical content, neutral over random: Llama 1B & 0.013 & 0.822 \\
Identical content, neutral over other replacement: Llama 1B & 0.002 & 0.176 \\
\bottomrule
\end{tabular}
\end{table*}
\FloatBarrier

\section{Depth, Row Coverage and Conditional Scoring versus Generation}
\label{app:evidence-depth}

The descriptive answer scan reuses the original 35 blocks, nine interfaces and fixed
source weights from 50\% depth. Span-aligned addition copies the same replacement's entity rows
tokenwise, interpolates later question and cue rows and copies candidate-prefix rows.
With token IDs matched as in Section~\ref{sec:answer-depth}, it reproduces the edit at
the decoder input by construction. Tables~\ref{tab:round11-depth-default}
and~\ref{tab:round11-rows} report the default-encoding depth and source/row comparisons;
Qwen2.5-7B lacks the target-relation row subsets and neutral-source runs. The nine relative interfaces are
$0,1/8,1/4,3/8,1/2,5/8,3/4,7/8,1$, rounded to the nearest layer. Here $d=0$ is
the effective decoder input and $d=1$ the last layer's output before final normalization.
Continuation still applies final normalization and the released head, including Gemma's
logit soft-capping, so hidden and direct reuse need not agree at $d=1$
(Table~\ref{tab:round11-depth-default}).
Entity rows (E) cover the minimal changed span; post-entity question rows (Q),
the answer cue (C), and candidate-prefix rows (A) partition the later suffix.
EQ retains entity, question and cue rows but omits answer-prefix rows.
Row-subset patches are zero elsewhere. Oracle patches use the target's own
edited-minus-clean states on those same rows.

Every unpatched continuation must reproduce clean logits within $10^{-3}$.
The separate cross-run tolerance is $10^{-4}$ in token log probability: the
Qwen2.5-7B depth re-run differs by $1.05\times10^{-4}$ in one of 3,192
construction/candidate comparisons after changing the float32 batch shape from
17 to 11 rows. Its curve therefore uses the earlier saved plain-format scan;
the new target-relation row subsets and neutral-source runs are not reported.

\begin{table*}[!htbp]
\centering
\scriptsize
\caption{Answer reuse across depth in each checkpoint's default encoding: percentage of the zero-prediction error removed on the 35 original blocks by exact span alignment with other-relation sources (all, entity or post-entity rows), by target-relation sources (all rows), and by the target's own edited-minus-clean change on entity or post-entity rows. Qwen 7B uses its saved plain-format records from the earlier scan, since its depth re-run failed the consistency check. The last column is direct reuse, which does not depend on depth. The table includes the plotted series in Figure~\ref{fig:answer-depth-exact} and additional row subsets.}
\label{tab:round11-depth-default}
\setlength{\tabcolsep}{3pt}
\begin{tabular}{@{}llrrrrrrrrrr@{}}
\toprule
Checkpoint & Construction & 0 & 1/8 & 1/4 & 3/8 & 1/2 & 5/8 & 3/4 & 7/8 & 1 & Direct \\
\midrule
Qwen 0.5B & Exact, all rows & 100.0 & 77.2 & 66.2 & 56.2 & 52.7 & 45.2 & 35.1 & 21.9 & 1.9 & $-$1.9 \\
 & Exact, entity rows & 100.0 & 75.5 & 47.3 & 37.3 & 35.9 & 26.9 & 9.6 & 5.0 & 0.0 &  \\
 & Exact, post-entity rows & 0.0 & $-$1.4 & 3.1 & 3.1 & $-$2.4 & $-$4.5 & 17.8 & 19.7 & 1.9 &  \\
 & Target relation, all rows & 100.0 & 81.2 & 72.1 & 69.8 & 62.4 & 57.3 & 52.3 & 47.6 & 45.1 &  \\
 & Own entity-row change & 100.0 & 85.0 & 54.9 & 48.2 & 40.9 & 25.7 & 8.6 & 4.2 & 0.0 &  \\
 & Own post-entity change & 0.0 & 1.0 & 11.1 & 10.3 & 7.0 & 15.7 & 41.3 & 63.7 & 100.0 &  \\
\midrule
Llama 1B & Exact, all rows & 100.0 & 84.0 & 72.9 & 68.5 & 63.3 & 55.0 & 52.5 & 41.6 & 34.4 & 32.4 \\
 & Exact, entity rows & 100.0 & 82.3 & 59.3 & 48.7 & 40.7 & 17.7 & 13.8 & 12.1 & 0.0 &  \\
 & Exact, post-entity rows & 0.0 & $-$0.2 & 7.8 & 9.5 & 10.4 & 17.5 & 24.9 & 28.8 & 34.4 &  \\
 & Target relation, all rows & 100.0 & 87.2 & 80.5 & 77.2 & 75.4 & 71.7 & 69.4 & 70.3 & 64.5 &  \\
 & Own entity-row change & 100.0 & 88.4 & 67.7 & 51.6 & 42.4 & 17.7 & 12.8 & 11.6 & 0.0 &  \\
 & Own post-entity change & 0.0 & 0.9 & 10.0 & 13.9 & 17.4 & 36.1 & 48.5 & 62.4 & 100.0 &  \\
\midrule
Gemma 2B & Exact, all rows & 100.0 & 87.4 & 80.8 & 72.3 & 69.8 & 63.5 & 50.0 & 39.9 & 30.7 & 29.9 \\
 & Exact, entity rows & 100.0 & 84.9 & 75.8 & 56.9 & 52.9 & 40.0 & 19.8 & 8.5 & 0.0 &  \\
 & Exact, post-entity rows & 0.0 & 3.3 & 3.8 & 10.9 & 14.6 & 19.6 & 31.9 & 38.7 & 30.7 &  \\
 & Target relation, all rows & 100.0 & 90.4 & 86.2 & 80.5 & 79.1 & 75.0 & 69.7 & 71.8 & 71.1 &  \\
 & Own entity-row change & 100.0 & 91.0 & 79.5 & 61.7 & 57.4 & 42.6 & 20.9 & 8.8 & 0.0 &  \\
 & Own post-entity change & 0.0 & 3.7 & 4.7 & 18.8 & 22.4 & 38.7 & 67.3 & 90.4 & 100.0 &  \\
\midrule
Qwen 7B & Exact, all rows & 100.0 & 77.2 & 68.2 & 58.1 & 56.9 & 41.3 & 26.3 & $-$0.7 & $-$23.5 & $-$27.9 \\
 & Exact, entity rows & 100.0 & 69.7 & 62.3 & 40.3 & 37.7 & 28.0 & 18.6 & $-$3.1 & 0.0 &  \\
 & Exact, post-entity rows & 0.0 & $-$0.9 & $-$0.2 & 8.9 & 15.4 & $-$1.6 & $-$4.2 & $-$3.1 & $-$23.5 &  \\
 & Target relation, all rows & 100.0 & 81.6 & 74.3 & 63.7 & 60.9 & 56.3 & 56.9 & 46.8 & 42.8 &  \\
 & Own entity-row change & 100.0 & 79.5 & 72.7 & 45.9 & 44.2 & 31.7 & 21.1 & $-$3.1 & 0.0 &  \\
 & Own post-entity change & 0.0 & $-$0.2 & $-$0.3 & 24.0 & 28.2 & 30.0 & 43.3 & 82.9 & 100.0 &  \\
\midrule
Llama 8B & Exact, all rows & 100.0 & 83.5 & 74.8 & 66.8 & 46.4 & 30.3 & 13.9 & $-$14.6 & $-$18.2 & $-$14.2 \\
 & Exact, entity rows & 100.0 & 77.4 & 61.5 & 51.7 & 28.3 & 1.3 & $-$1.8 & $-$0.8 & 0.0 &  \\
 & Exact, post-entity rows & 0.0 & 3.4 & 7.3 & 12.9 & 11.9 & 19.0 & 9.4 & $-$17.0 & $-$18.2 &  \\
 & Target relation, all rows & 100.0 & 86.7 & 83.2 & 78.7 & 72.2 & 65.1 & 63.2 & 60.8 & 58.5 &  \\
 & Own entity-row change & 100.0 & 81.9 & 68.4 & 54.9 & 32.1 & 2.1 & $-$1.0 & $-$0.7 & 0.0 &  \\
 & Own post-entity change & 0.0 & 4.4 & 9.4 & 14.0 & 34.1 & 69.5 & 81.4 & 90.0 & 100.0 &  \\
\midrule
Gemma 9B & Exact, all rows & 100.0 & 91.5 & 81.3 & 78.2 & 68.7 & 56.1 & 34.1 & 13.3 & $-$11.8 & $-$23.3 \\
 & Exact, entity rows & 100.0 & 88.5 & 76.3 & 69.7 & 53.9 & 37.0 & 14.8 & 10.1 & 0.0 &  \\
 & Exact, post-entity rows & 0.0 & 0.0 & 2.3 & 4.5 & 9.3 & 21.4 & 29.1 & 13.4 & $-$11.8 &  \\
 & Target relation, all rows & 100.0 & 93.1 & 87.2 & 83.7 & 78.2 & 75.0 & 67.6 & 68.5 & 66.4 &  \\
 & Own entity-row change & 100.0 & 93.3 & 80.9 & 72.4 & 58.0 & 39.3 & 16.1 & 11.2 & 0.0 &  \\
 & Own post-entity change & 0.0 & $-$0.8 & 4.1 & 6.2 & 13.6 & 48.4 & 77.1 & 86.3 & 100.0 &  \\
\bottomrule
\end{tabular}
\end{table*}
\begin{table*}[!htbp]
\centering
\footnotesize
\caption{Row subsets by source relation at $d=1/4$ and $d=1/2$ in the default encoding (35 original blocks): percentage of the zero-prediction error removed by exact alignment on all rows, entity, question and cue rows (EQ), entity rows (E) or post-entity rows (QA) with other-relation, target-relation or relation-neutral sources (neutral entity rows with equal weights), and by the target's own entity-row change. The last column is the target-relation all-row gain over entity rows (nats, pointwise 95\% interval). Qwen 7B uses its saved plain-format records from the earlier scan, which lack the target-relation subsets and neutral sources.}
\label{tab:round11-rows}
\setlength{\tabcolsep}{2.5pt}
\begin{tabular}{@{}llrrrrrrrrrrr@{}}
\toprule
Checkpoint & $d$ & Oth. all & Oth. EQ & Oth. E & Oth. QA & Tgt. all & Tgt. E & Tgt. QA & Neu. all & Neu. E & Own E & All over E \\
\midrule
Qwen 0.5B & 1/4 & 66.2 & 49.8 & 47.3 & 3.1 & 72.1 & 49.8 & 6.6 & 60.9 & 47.9 & 54.9 & 0.64 [0.34, 1.02] \\
 & 1/2 & 52.7 & 36.5 & 35.9 & $-$2.4 & 62.4 & 35.3 & $-$0.6 & 45.1 & 35.1 & 40.9 & 0.78 [0.39, 1.20] \\
Llama 1B & 1/4 & 72.9 & 59.3 & 59.3 & 7.8 & 80.5 & 63.1 & 8.3 & 72.8 & 60.7 & 67.7 & 0.49 [0.28, 0.74] \\
 & 1/2 & 63.3 & 43.3 & 40.7 & 10.4 & 75.4 & 40.2 & 15.4 & 54.1 & 39.4 & 42.4 & 0.99 [0.61, 1.41] \\
Gemma 2B & 1/4 & 80.8 & 76.9 & 75.8 & 3.8 & 86.2 & 76.0 & 4.8 & 81.3 & 75.4 & 79.5 & 0.44 [0.27, 0.62] \\
 & 1/2 & 69.8 & 58.7 & 52.9 & 14.6 & 79.1 & 53.5 & 17.4 & 70.4 & 51.4 & 57.4 & 1.10 [0.78, 1.45] \\
Qwen 7B & 1/4 & 68.2 & 63.9 & 62.3 & $-$0.2 & 74.3 & -- & -- & -- & -- & 72.7 & -- \\
 & 1/2 & 56.9 & 41.9 & 37.7 & 15.4 & 60.9 & -- & -- & -- & -- & 44.2 & -- \\
Llama 8B & 1/4 & 74.8 & 65.3 & 61.5 & 7.3 & 83.2 & 65.6 & 6.7 & 75.8 & 64.0 & 68.4 & 0.74 [0.48, 1.01] \\
 & 1/2 & 46.4 & 30.7 & 28.3 & 11.9 & 72.2 & 32.4 & 29.5 & 52.1 & 28.3 & 32.1 & 1.67 [1.19, 2.22] \\
Gemma 9B & 1/4 & 81.3 & 79.2 & 76.3 & 2.3 & 87.2 & 77.5 & 4.8 & 84.4 & 76.6 & 80.9 & 0.45 [0.29, 0.61] \\
 & 1/2 & 68.7 & 58.5 & 53.9 & 9.3 & 78.2 & 55.9 & 11.8 & 70.9 & 53.3 & 58.0 & 1.03 [0.71, 1.38] \\
\bottomrule
\end{tabular}
\end{table*}

\paragraph{Combining row responses.}
\label{app:row-additivity}
On the original 35 blocks in default encoding at $d=1/2$, adding the target's separately continued entity-row and post-entity-row predictions as contrast changes removes 51.7, 55.3, 68.5, 54.5, 48.7, 61.9\% of no-change error, in Qwen 0.5B, Llama 1B, Gemma 2B, Qwen 7B, Llama 8B and Gemma 9B order. Patching both row sets together instead removes 100\% by reproducing the edit. These are errors evaluated after adding the predicted contrast changes, not sums of the two individual error-removal shares; absolute error is nonlinear.

Table~\ref{tab:round11-readouts} evaluates the full-vocabulary distribution at the answer
cue, without choosing a candidate prefix. Table~\ref{tab:round11-generation} instead
lets greedy generation follow its own tokens, using one question-plus-cue capture.
These results are not interchangeable with per-prefix candidate scoring: capture,
construction and evaluation scope differ, so the gap does not isolate autoregressive
error accumulation. Low generation fidelity is reported alongside the stronger
conditional-score results.

\begin{table*}[!htbp]
\centering
\footnotesize
\caption{Full-vocabulary readouts at the answer cue, whose next-token distribution does not depend on the candidate set: KL recovery, one minus the mean block KL divergence from the edited target over that of the clean target (\%), and agreement of the most probable token with the edited target's (\%), over the 99 blocks of three cohorts in each checkpoint's default encoding. Constructions use other-relation sources unless marked target (Tgt.); one vector (Vec.) at native scale. Interp.: suffix interpolation; Span: span-aligned addition; Repl.: entity-state replacement; Dir.: direct output reuse; E: entity rows; later: every row from the entity onward. The clean target's top-token agreement is the baseline.}
\label{tab:round11-readouts}
\setlength{\tabcolsep}{3.5pt}
\begin{tabular}{@{}lrrrrrrrrrr@{}}
\toprule
 & Interp. & Span all & Span E & Repl. E & Vec. E & Vec. later & Tgt. all & Dir. oth. & Dir. tgt. & Clean \\
\midrule
\multicolumn{11}{@{}l}{\emph{KL recovery}} \\
Qwen 0.5B & 88 & 91 & 89 & 89 & 42 & $-$42 & 92 & 44 & 80 & -- \\
Llama 1B & 97 & 98 & 97 & 97 & 25 & 6 & 98 & 87 & 93 & -- \\
Gemma 2B & 94 & 95 & 93 & 93 & 45 & 14 & 97 & 67 & 88 & -- \\
Qwen 7B & 94 & 95 & 94 & 94 & 32 & 33 & 96 & 76 & 88 & -- \\
Llama 8B & 94 & 95 & 91 & 92 & 38 & 36 & 97 & 72 & 90 & -- \\
Gemma 9B & 95 & 97 & 94 & 94 & 52 & 51 & 98 & 69 & 90 & -- \\
\midrule
\multicolumn{11}{@{}l}{\emph{Top-token agreement}} \\
Qwen 0.5B & 75 & 77 & 76 & 76 & 52 & 30 & 81 & 46 & 63 & 37 \\
Llama 1B & 88 & 89 & 91 & 89 & 32 & 25 & 90 & 64 & 70 & 20 \\
Gemma 2B & 77 & 78 & 75 & 75 & 55 & 26 & 84 & 50 & 69 & 44 \\
Qwen 7B & 76 & 78 & 74 & 75 & 21 & 15 & 77 & 45 & 56 & 13 \\
Llama 8B & 85 & 86 & 80 & 80 & 32 & 30 & 87 & 51 & 69 & 19 \\
Gemma 9B & 84 & 85 & 82 & 84 & 39 & 31 & 89 & 51 & 66 & 22 \\
\bottomrule
\end{tabular}
\end{table*}
\begin{table*}[!htbp]
\centering
\footnotesize
\caption{Greedy generation of up to 16 tokens after the answer cue, over 396 targets of three cohorts in each checkpoint's default encoding. Changes are added at the 1/2 interface: span-aligned entity, question and cue rows from one capture (Span, other or target relation), replaced entity states (Repl.), or one vector at every row from the entity on, generated rows included (Vec.). Changed: the edited target's answer string, the text before the first line break, differs from the clean target's (\%). Exact: answer string identical to the edited target's (\%). In set: greedy answer equal to a scored candidate (\%). Contains (post hoc): answer contains a published target-relation answer of the replacement, for the edited target (\% of targets, count) and, among those targets, for the Span oth. and Vec. predictions and for the clean target (\%).}
\label{tab:round11-generation}
\setlength{\tabcolsep}{3.5pt}
\begin{tabular}{@{}lrrrrrrrrrrr@{}}
\toprule
 & & \multicolumn{4}{c}{Exact} & \multicolumn{2}{c}{In set} & \multicolumn{4}{c}{Contains} \\
\cmidrule(lr){3-6}\cmidrule(lr){7-8}\cmidrule(lr){9-12}
Checkpoint & Changed & Span oth. & Span tgt. & Repl. & Vec. & Clean & Edited & Edited & Span oth. & Vec. & Clean \\
\midrule
Qwen 0.5B & 96 & 18 & 18 & 17 & 2 & 7 & 6 & 11 (42) & 81 & 29 & 31 \\
Llama 1B & 100 & 5 & 5 & 5 & 0 & 0 & 0 & 15 (60) & 28 & 32 & 27 \\
Gemma 2B & 96 & 17 & 21 & 14 & 3 & 6 & 7 & 21 (82) & 66 & 29 & 21 \\
Qwen 7B & 98 & 6 & 7 & 5 & 1 & 4 & 4 & 18 (72) & 43 & 28 & 14 \\
Llama 8B & 98 & 10 & 11 & 8 & 4 & 7 & 9 & 39 (154) & 46 & 38 & 19 \\
Gemma 9B & 97 & 23 & 26 & 22 & 8 & 9 & 11 & 31 (123) & 67 & 41 & 18 \\
\bottomrule
\end{tabular}
\end{table*}
\FloatBarrier

\section{Standard Patches and the Single-Vector Comparison}
\label{app:evidence-constructions}

The default-encoding comparison over 99 blocks favors span-aligned addition, by 0.12--0.28 nats with all
six pointwise block-bootstrap intervals excluding zero
(Table~\ref{tab:core-extra-comparisons}c).
Coverage and construction change together, so that comparison does not isolate post-entity rows.

Table~\ref{tab:round11-families} reports the testbed families, including the undetected
entity-addition versus state-replacement differences and the family-specific knockout
outcomes. The stronger single-vector extension selects among 90 constructions on the
original cohort and evaluates on the two already-studied fresh cohorts; the selection
and evaluation sets are distinct but neither is an unseen research setting. Directions,
placements, averaging and amplitudes vary, so the comparison is between tested
constructions, not an isolated effect of row structure. Tables~\ref{tab:b1-families}
and~\ref{tab:b1-intervals} report all six checkpoints and all four families, including
failed and pooled-correction-sensitive tests.

\paragraph{The 90 constructions.}
The single-pair vector is the inverse-distance-weighted change at the two
other-relation sources' last entity row. Its four placements are the target's
last entity row, every entity row, every row from the entity onward, or answer
positions. Seven coefficients, $1/8,1/4,1/2,1,2,4,8$, give 28 constructions.
The sixteen-context vector averages last-entity-row changes with equal weights
over those two sources, the two neutral templates and twelve further relation-neutral
templates, giving another 28 at the same placements and coefficients.
The twelve templates are \emph{Here is a note about $X$.}, \emph{Describe $X$.},
\emph{Who or what is $X$?}, \emph{Write a few words about $X$.},
\emph{Give some facts about $X$.}, \emph{Summarize what is known about $X$.},
\emph{The topic is $X$.}, \emph{Consider $X$.}, \emph{Let us talk about $X$.},
\emph{I want to learn about $X$.}, \emph{Name one fact about $X$.} and
\emph{Have you heard of $X$?}, rendered with the same cue and candidate prefix.
Neither vector uses target-relation questions. Taking either vector at the answer
cue instead and adding it at answer positions or every later row gives 28 more.
The final six use either entity-vector direction at entity rows, every later row
or answer positions, with each row scaled to the span-aligned response's norm.
All continue through the target's remaining layers. The best construction
minimizes mean block answer error on the original 35 blocks, with exact ties
resolved by the fixed construction order, and is evaluated on the other 64
already-studied blocks. The norm-matched constructions borrow magnitude
information from span-aligned reuse; this is not a comparison at equal information.
The earlier D1c instead tunes the coefficient of one vector at each entity row.

\begin{table*}[!htbp]
\centering
\footnotesize
\caption{Confirmatory families of the testbed and read-out protocol, each six one-sided (D1a, D2a, D2b) or two-sided (D1b, D1c) exact sign tests with Holm correction within the family, in each checkpoint's default encoding. Cells give positive / negative block counts, the mean signed error difference in nats and the Holm-adjusted $p$. D1a: later-row vector error minus span-aligned error; D1b: state-replacement error minus additive entity-row error on the same rows; D1c: tuned-vector error minus span-aligned error, with tuning on the original cohort (fresh cohorts, 64 blocks); D2a: zero minus neutral selective error for identical entity-row changes; D2b: error with answer-position attention to entity rows blocked minus error with attention to other prompt rows blocked, sparing the first token and final cue row in the latter. Other families use 99 blocks. A dagger marks tests that do not pass the final 223-test pooled Holm correction (Table~\ref{tab:round10-pooled}). For 80\% power the true share of favoring blocks must reach 0.63--0.66 in the one-sided families, 0.64--0.68 in D1b and 0.68--0.71 in D1c (last to first Holm rank).}
\label{tab:round11-families}
\setlength{\tabcolsep}{4pt}
\begin{tabular}{@{}llllll@{}}
\toprule
Checkpoint & D1a & D1b & D1c & D2a & D2b \\
\midrule
Qwen 0.5B & 99/0, 7.61, $<$0.001 & 55/44, $-$0.06, 1.000 & 50/14, 0.85, $<$0.001 & 71/28, 0.40, $<$0.001 & 41/58, 0.10, 1.000 \\
Llama 1B & 97/2, 6.76, $<$0.001 & 50/49, $-$0.01, 1.000 & 51/13, 0.67, $<$0.001 & 60/39, 0.16, 0.022$^\dagger$ & 72/27, 0.92, $<$0.001 \\
Gemma 2B & 99/0, 6.50, $<$0.001 & 43/56, $-$0.06, 1.000 & 51/13, 0.77, $<$0.001 & 78/21, 0.75, $<$0.001 & 46/53, $-$0.03, 1.000 \\
Qwen 7B & 97/2, 7.26, $<$0.001 & 51/48, 0.01, 1.000 & 42/22, 1.06, 0.034$^\dagger$ & 68/31, 1.69, $<$0.001 & 45/54, $-$0.11, 1.000 \\
Llama 8B & 95/4, 6.04, $<$0.001 & 43/56, $-$0.05, 1.000 & 42/22, 0.87, 0.034$^\dagger$ & 69/30, 0.74, $<$0.001 & 76/23, 1.39, $<$0.001 \\
Gemma 9B & 96/3, 5.26, $<$0.001 & 56/43, 0.01, 1.000 & 52/12, 1.09, $<$0.001 & 83/16, 1.29, $<$0.001 & 34/65, $-$0.62, 1.000 \\
\bottomrule
\end{tabular}
\end{table*}
\begin{table*}[!htbp]
\centering
\footnotesize
\caption{Confirmatory families of the steering-completion protocol, each six exact sign tests with Holm correction within the family, in each checkpoint's default encoding. Cells give positive / negative block counts, the mean signed error difference in nats and the Holm-adjusted $p$. E1 (one-sided, fresh cohorts, 64 blocks): the single-pair vector at every later row, at the coefficient chosen on the original cohort (Qwen 0.5B 1/8, Llama 1B 1/4, Gemma 2B 1/2, Qwen 7B 1/2, Llama 8B 1/2, Gemma 9B 1/2), has lower error than zero prediction. E2 (two-sided, 64 blocks): best-vector error minus span-aligned error, for the best of the 90 single-vector constructions chosen on the original cohort (Qwen 0.5B, single-pair vector at each entity row, coefficient 1; Llama 1B, norm-matched single-pair direction at the entity rows; Gemma 2B, norm-matched single-pair direction at every later row; Qwen 7B, norm-matched sixteen-context direction at every later row; Llama 8B, norm-matched sixteen-context direction at every later row; Gemma 9B, single-pair vector at each entity row, coefficient 1); for Qwen 0.5B and Gemma 9B this is the vector tuned for D1c, so their E2 tests repeat D1c on the same blocks (Table~\ref{tab:round11-families}). E3 (two-sided, 99 blocks): single-pair error minus sixteen-context entity-vector error, both at each entity row at coefficient 1. E4 (one-sided, 99 blocks): the norm-matched single-pair direction at every later row has lower error than zero prediction. A dagger marks tests that do not pass the final 223-test pooled Holm correction (Table~\ref{tab:round10-pooled}). For 80\% power the true share of favoring blocks must reach 0.67--0.71 in E1, 0.68--0.71 in E2, 0.64--0.68 in E3 and 0.63--0.66 in E4 (last to first Holm rank).}
\label{tab:b1-families}
\setlength{\tabcolsep}{4pt}
\begin{tabular}{@{}lllll@{}}
\toprule
Checkpoint & E1 & E2 & E3 & E4 \\
\midrule
Qwen 0.5B & 38/26, 0.13, 0.253 & 50/14, 0.85, $<$0.001 & 42/57, $-$0.08, 0.954 & 48/51, 0.20, 0.656 \\
Llama 1B & 44/20, 0.45, 0.007$^\dagger$ & 51/13, 0.66, $<$0.001 & 47/52, $-$0.04, 1.000 & 66/33, 0.54, 0.001$^\dagger$ \\
Gemma 2B & 37/27, 0.53, 0.253 & 51/13, 0.82, $<$0.001 & 56/43, 0.04, 1.000 & 76/23, 1.65, $<$0.001 \\
Qwen 7B & 38/26, 0.97, 0.253 & 45/19, 1.38, 0.003$^\dagger$ & 51/48, 0.07, 1.000 & 72/27, 2.12, $<$0.001 \\
Llama 8B & 45/19, 1.47, 0.004$^\dagger$ & 38/26, 0.48, 0.169 & 46/53, $-$0.03, 1.000 & 81/18, 1.79, $<$0.001 \\
Gemma 9B & 49/15, 2.05, $<$0.001 & 52/12, 1.09, $<$0.001 & 47/52, $-$0.06, 1.000 & 81/18, 2.21, $<$0.001 \\
\bottomrule
\end{tabular}
\end{table*}
\begin{table*}[!htbp]
\centering
\footnotesize
\caption{Mean signed error differences defined in Table~\ref{tab:b1-families} (nats) with 95\% intervals resampling blocks within cohorts (Within) or relation labels, each block weighted by the product of its two labels' counts (Labels; Appendix~\ref{app:statistical-details}). The lower panel gives the share of blocks with positive differences and its label-resampled interval; $\dagger$ marks intervals that include one half.}
\label{tab:b1-intervals}
\setlength{\tabcolsep}{1.5pt}
\begin{tabular}{@{}lrrrrrrrr@{}}
\toprule
 & \multicolumn{2}{c}{E1} & \multicolumn{2}{c}{E2} & \multicolumn{2}{c}{E3} & \multicolumn{2}{c}{E4} \\
\cmidrule(lr){2-3}\cmidrule(lr){4-5}\cmidrule(lr){6-7}\cmidrule(lr){8-9}
Checkpoint & Within & Labels & Within & Labels & Within & Labels & Within & Labels \\
\midrule
Qwen 0.5B & [0.00, 0.27] & [$-$0.10, 0.37] & [0.53, 1.23] & [0.33, 1.62] & [$-$0.15, $-$0.02] & [$-$0.19, 0.03] & [$-$0.08, 0.50] & [$-$0.31, 0.79] \\
Llama 1B & [0.21, 0.69] & [0.08, 0.87] & [0.45, 0.88] & [0.35, 1.05] & [$-$0.10, 0.01] & [$-$0.14, 0.04] & [0.27, 0.81] & [0.12, 0.99] \\
Gemma 2B & [0.01, 1.05] & [$-$0.34, 1.41] & [0.57, 1.07] & [0.41, 1.25] & [$-$0.03, 0.12] & [$-$0.09, 0.18] & [1.24, 2.08] & [0.93, 2.49] \\
Qwen 7B & [0.13, 1.80] & [$-$0.53, 2.45] & [0.69, 2.05] & [0.18, 2.50] & [$-$0.11, 0.24] & [$-$0.24, 0.37] & [1.55, 2.70] & [1.15, 3.16] \\
Llama 8B & [0.88, 2.10] & [0.46, 2.57] & [0.16, 0.80] & [$-$0.11, 1.10] & [$-$0.09, 0.02] & [$-$0.13, 0.05] & [1.35, 2.24] & [1.00, 2.66] \\
Gemma 9B & [1.45, 2.67] & [1.05, 3.04] & [0.74, 1.47] & [0.50, 1.77] & [$-$0.13, 0.02] & [$-$0.22, 0.08] & [1.74, 2.70] & [1.37, 3.06] \\
\midrule
\multicolumn{9}{@{}l}{\emph{Share of blocks with positive differences, labels resampled}} \\
Qwen 0.5B & \multicolumn{2}{c}{0.59 [0.38, 0.79]$^\dagger$} & \multicolumn{2}{c}{0.78 [0.60, 0.93]} & \multicolumn{2}{c}{0.42 [0.27, 0.59]$^\dagger$} & \multicolumn{2}{c}{0.48 [0.30, 0.67]$^\dagger$} \\
Llama 1B & \multicolumn{2}{c}{0.69 [0.47, 0.88]$^\dagger$} & \multicolumn{2}{c}{0.80 [0.62, 0.94]} & \multicolumn{2}{c}{0.47 [0.32, 0.64]$^\dagger$} & \multicolumn{2}{c}{0.67 [0.51, 0.81]} \\
Gemma 2B & \multicolumn{2}{c}{0.58 [0.36, 0.78]$^\dagger$} & \multicolumn{2}{c}{0.80 [0.60, 0.95]} & \multicolumn{2}{c}{0.57 [0.42, 0.71]$^\dagger$} & \multicolumn{2}{c}{0.77 [0.63, 0.89]} \\
Qwen 7B & \multicolumn{2}{c}{0.59 [0.39, 0.78]$^\dagger$} & \multicolumn{2}{c}{0.70 [0.52, 0.87]} & \multicolumn{2}{c}{0.52 [0.34, 0.68]$^\dagger$} & \multicolumn{2}{c}{0.73 [0.57, 0.86]} \\
Llama 8B & \multicolumn{2}{c}{0.70 [0.49, 0.88]$^\dagger$} & \multicolumn{2}{c}{0.59 [0.38, 0.80]$^\dagger$} & \multicolumn{2}{c}{0.46 [0.30, 0.63]$^\dagger$} & \multicolumn{2}{c}{0.82 [0.66, 0.94]} \\
Gemma 9B & \multicolumn{2}{c}{0.77 [0.55, 0.93]} & \multicolumn{2}{c}{0.81 [0.62, 0.96]} & \multicolumn{2}{c}{0.47 [0.30, 0.66]$^\dagger$} & \multicolumn{2}{c}{0.82 [0.68, 0.93]} \\
\bottomrule
\end{tabular}
\end{table*}
\FloatBarrier

\section{Receiving-Question Dependence and Content Controls}
\label{app:evidence-content}

The knockouts block attention from the answer positions above the interface.
KO-E targets entity rows; KO-PQ targets other prompt rows except the first token and
the final cue row. Table~\ref{tab:round11-knockouts} reports their family-dependent
outcomes and the shift of the clean answer contrast caused by KO-E. These are route
interventions, not identified heads or features, and their clean-reference changes
limit a mechanistic interpretation.

\begin{table*}[!htbp]
\centering
\footnotesize
\caption{Attention knockouts above the interface: share of the zero-prediction error removed under each knockout divided by the share without it, for other-relation entity rows / identical neutral entity rows / the target's own entity-row change, over the 99 blocks of three cohorts. Knockouts block attention from the answer positions (final cue row and candidate prefix) to entity rows (E), to the question and cue rows between entity and final cue row (Q), to rows before the entity except the first (P), or to both (PQ), in every remaining layer or in its first or second half; each prediction is measured from the clean target under the same knockout. Blocking attention from every post-entity row to the entity rows removes the entity-row effect entirely, as the causal structure requires; this check is not tabulated. Shift: mean absolute change of the clean target's answer contrast under KO-E (nats).}
\label{tab:round11-knockouts}
\setlength{\tabcolsep}{3pt}
\begin{tabular}{@{}llllllll@{}}
\toprule
Checkpoint & KO-E & KO-E first half & KO-E second half & KO-Q & KO-P & KO-PQ & Shift \\
\midrule
Qwen 0.5B & 0.03/0.03/0.03 & 0.50/0.51/0.45 & 0.60/0.62/0.59 & 0.77/0.78/0.77 & 0.65/0.60/0.64 & 0.14/$-$0.09/0.15 & 1.72 \\
Llama 1B & 0.12/0.12/0.11 & 0.42/0.44/0.42 & 0.73/0.71/0.72 & 0.97/0.96/0.96 & 0.94/0.93/0.94 & 0.91/0.90/0.90 & 1.21 \\
Gemma 2B & 0.55/0.54/0.55 & 0.69/0.69/0.67 & 0.90/0.90/0.90 & 0.61/0.61/0.60 & 1.00/1.00/0.99 & 0.54/0.53/0.53 & 1.07 \\
Qwen 7B & 0.15/0.12/0.11 & 0.66/0.62/0.58 & 0.56/0.59/0.56 & 0.58/0.59/0.59 & 0.74/0.75/0.74 & 0.11/0.16/0.23 & 3.89 \\
Llama 8B & 0.11/0.12/0.12 & 0.14/0.16/0.15 & 0.95/0.93/0.93 & 1.01/1.00/1.00 & 0.97/0.96/0.97 & 0.93/0.92/0.93 & 1.38 \\
Gemma 9B & 0.70/0.71/0.69 & 0.78/0.78/0.78 & 0.95/0.95/0.94 & 0.69/0.67/0.67 & 0.99/1.00/0.99 & 0.47/0.46/0.48 & 0.98 \\
\bottomrule
\end{tabular}
\end{table*}

For the identical-content experiment, the two receiving questions get identical
relation-neutral entity-row changes. The controls use either the other replacement's
neutral change or eight random perturbations with matched row norms. The original
two test families remain in Table~\ref{tab:controls-families}.
NEU averages the changes from \emph{Tell me about $X$.} and \emph{What do you know about $X$?}
with equal weights. DR maps the other replacement's neutral change to the target's
entity span by relative-position interpolation, copying tokenwise when lengths
agree; DRN rescales each DR row to NEU's norm. RND uses independent Gaussian rows
with NEU's row norms, averaging selective errors over eight seeded draws.
Both relation-swapped targets receive the same injected content for a given
replacement. G1 tests NEU against RND and G2 against DR.
Tables~\ref{tab:controls-reductions} and~\ref{tab:controls-magnitudes} report absolute
selective-error reductions, random-control sensitivity and predicted magnitudes:
random changes are not universally ineffective, and generic route attenuation does not
by itself establish content specificity. The same-replacement change outperforms the
controls within the specified families, with pooled-correction exceptions in
Table~\ref{tab:round10-pooled}. Mean/share sensitivity intervals for the control
comparisons are in Table~\ref{tab:controls-intervals}; those for neutral content
against zero are in Table~\ref{tab:core-local-sensitivity}b.
On the 28--32 blocks per checkpoint where DR needs no span interpolation,
NEU beats DR in 18--25 blocks; its uncorrected sign-test $p$ exceeds 0.05 only
on Qwen2.5-7B ($p=0.181$). For the neutral
content itself against zero, Llama-3.2-1B's label-resampled mean interval is
[$-0.03$, 0.37] nats, despite its original within-family pass
(Table~\ref{tab:core-local-sensitivity}b).

\begin{table}[!htbp]
\centering
\footnotesize
\caption{Confirmatory families of the identical-content controls, each six one-sided exact sign tests over the 99 blocks with Holm correction within the family, in each checkpoint's default encoding. Cells give blocks favoring / opposing the hypothesis, the mean block difference of selective error in nats (positive favors it) and the Holm-adjusted $p$. G1: the same replacement's identical neutral entity-row change has lower selective error than identical random entity-row perturbations with its row norms, averaged over eight draws. G2: it has lower selective error than the identical neutral change of the block's other replacement mapped onto the target's entity span. A dagger marks tests that do not pass when all 223 confirmatory tests form one Holm family (pooled $p$: G1 Llama 1B 0.822; G2 Llama 1B 0.176). For 80\% power the true share of favoring blocks must reach 0.63--0.66 (last to first Holm rank).}
\label{tab:controls-families}
\setlength{\tabcolsep}{4pt}
\begin{tabular}{@{}lll@{}}
\toprule
Checkpoint & G1 & G2 \\
\midrule
Qwen 0.5B & 74/25, 0.42, $<$0.001 & 71/28, 0.46, $<$0.001 \\
Llama 1B & 61/38, 0.15, 0.013$^\dagger$ & 64/35, 0.21, 0.002$^\dagger$ \\
Gemma 2B & 81/18, 0.70, $<$0.001 & 75/24, 0.77, $<$0.001 \\
Qwen 7B & 75/24, 1.54, $<$0.001 & 68/31, 1.47, $<$0.001 \\
Llama 8B & 70/29, 0.67, $<$0.001 & 68/31, 0.64, $<$0.001 \\
Gemma 9B & 81/18, 1.25, $<$0.001 & 84/15, 1.19, $<$0.001 \\
\bottomrule
\end{tabular}
\end{table}
\begin{table*}[!htbp]
\centering
\footnotesize
\caption{Mean block differences of the two control families (nats; positive favors the hypothesis) with 95\% intervals resampling blocks within cohorts (Within) or relation labels, each block weighted by the product of its two labels' counts (Labels; Appendix~\ref{app:statistical-details}), and the share of favoring blocks with its label-resampled interval (Fav.); $\dagger$ marks share intervals that include one half.}
\label{tab:controls-intervals}
\setlength{\tabcolsep}{4pt}
\begin{tabular}{@{}lrrrrrr@{}}
\toprule
 & \multicolumn{3}{c}{G1} & \multicolumn{3}{c}{G2} \\
\cmidrule(lr){2-4}\cmidrule(lr){5-7}
Checkpoint & Within & Labels & Fav. & Within & Labels & Fav. \\
\midrule
Qwen 0.5B & [0.29, 0.57] & [0.17, 0.70] & 0.75 [0.59, 0.88] & [0.31, 0.62] & [0.17, 0.76] & 0.72 [0.52, 0.88] \\
Llama 1B & [0.05, 0.25] & [$-$0.02, 0.34] & 0.62 [0.44, 0.77]$^\dagger$ & [0.10, 0.33] & [0.01, 0.42] & 0.65 [0.48, 0.80]$^\dagger$ \\
Gemma 2B & [0.51, 0.90] & [0.36, 1.09] & 0.82 [0.66, 0.94] & [0.56, 1.00] & [0.40, 1.20] & 0.76 [0.59, 0.89] \\
Qwen 7B & [1.12, 1.96] & [0.76, 2.37] & 0.76 [0.58, 0.90] & [1.02, 1.93] & [0.67, 2.35] & 0.69 [0.51, 0.84] \\
Llama 8B & [0.44, 0.91] & [0.30, 1.10] & 0.71 [0.55, 0.85] & [0.43, 0.86] & [0.27, 1.06] & 0.69 [0.49, 0.86]$^\dagger$ \\
Gemma 9B & [0.94, 1.57] & [0.76, 1.83] & 0.82 [0.67, 0.94] & [0.92, 1.48] & [0.73, 1.73] & 0.85 [0.71, 0.95] \\
\bottomrule
\end{tabular}
\end{table*}
\begin{table*}[!htbp]
\centering
\footnotesize
\caption{Selective-error reduction from zero prediction, $|s|-|\hat s-s|$ (nats), of each identical entity-row construction over the 99 blocks, with 95\% intervals resampling blocks within cohorts. NEU: the same replacement's neutral change, the identical-content construction of Appendix~\ref{app:evidence-content}; DR: the other replacement's neutral change mapped onto the target's span; DRN: DR with each row rescaled to NEU's norm; RND: random rows with NEU's norms, averaged over eight draws, with its interval resampling relation labels (RND labels; Appendix~\ref{app:statistical-details}) and the range of the eight draws' means. $\ddagger$: the interval excludes zero although a bound rounds to 0.00.}
\label{tab:controls-reductions}
\setlength{\tabcolsep}{2.5pt}
\begin{tabular}{@{}lrrrrrr@{}}
\toprule
Checkpoint & NEU & DR & DRN & RND & RND labels & RND draws \\
\midrule
Qwen 0.5B & 0.40 [0.26, 0.56] & $-$0.06 [$-$0.17, 0.04] & $-$0.05 [$-$0.17, 0.06] & $-$0.02 [$-$0.09, 0.04] & [$-$0.15, 0.11] & $-$0.10 to 0.05 \\
Llama 1B & 0.16 [0.06, 0.27] & $-$0.05 [$-$0.12, 0.02] & $-$0.06 [$-$0.14, 0.02] & 0.01 [$-$0.02, 0.04] & [$-$0.04, 0.07] & $-$0.01 to 0.03 \\
Gemma 2B & 0.75 [0.54, 0.96] & $-$0.03 [$-$0.15, 0.09] & $-$0.02 [$-$0.13, 0.09] & 0.04 [0.00, 0.09] & [$-$0.04, 0.14] & 0.01 to 0.10 \\
Qwen 7B & 1.69 [1.21, 2.19] & 0.22 [$-$0.08, 0.54] & 0.31 [0.01, 0.63] & 0.15 [$-$0.08, 0.39] & [$-$0.21, 0.59] & 0.03 to 0.33 \\
Llama 8B & 0.74 [0.50, 1.00] & 0.11 [$-$0.02, 0.23] & 0.12 [$-$0.01, 0.24] & 0.07 [0.03, 0.12] & [0.01, 0.14] & 0.02 to 0.13 \\
Gemma 9B & 1.29 [0.98, 1.62] & 0.11 [$-$0.05, 0.27] & 0.11 [$-$0.06, 0.28] & 0.05 [0.00, 0.10]$^\ddagger$ & [$-$0.03, 0.14] & 0.02 to 0.07 \\
\bottomrule
\end{tabular}
\end{table*}
\begin{table*}[!htbp]
\centering
\footnotesize
\caption{Size of the predicted selective component and the full-metric share of the zero-prediction error removed by each identical entity-row construction over the 99 blocks: the mean observed $|s|$ (nats), each construction's mean $|\hat s|$ as a percentage of it (for RND, the mean over draws of each draw's mean $|\hat s|$), and the share removed on the complete answer contrast (\%). Constructions as in Table~\ref{tab:controls-reductions}.}
\label{tab:controls-magnitudes}
\setlength{\tabcolsep}{4pt}
\begin{tabular}{@{}lrrrrrrrrr@{}}
\toprule
 & & \multicolumn{4}{c}{$|\hat s|$ (\% of $|s|$)} & \multicolumn{4}{c}{Share removed (\%)} \\
\cmidrule(lr){3-6}\cmidrule(lr){7-10}
Checkpoint & $|s|$ & NEU & DR & DRN & RND & NEU & DR & DRN & RND \\
\midrule
Qwen 0.5B & 1.37 & 74.1 & 53.0 & 54.3 & 45.1 & 32.1 & $-$5.8 & $-$6.6 & $-$6.1 \\
Llama 1B & 0.89 & 80.5 & 48.9 & 52.6 & 31.1 & 44.5 & 0.3 & $-$0.8 & 3.3 \\
Gemma 2B & 1.76 & 74.7 & 38.2 & 39.7 & 21.4 & 50.7 & 0.3 & $-$0.1 & 3.5 \\
Qwen 7B & 4.48 & 76.4 & 43.0 & 43.3 & 37.6 & 37.6 & 5.1 & 5.2 & 4.9 \\
Llama 8B & 2.53 & 57.3 & 26.6 & 27.3 & 13.6 & 36.1 & 5.9 & 6.2 & 3.3 \\
Gemma 9B & 2.59 & 78.3 & 34.9 & 37.5 & 14.9 & 54.6 & 4.2 & 4.8 & 2.3 \\
\bottomrule
\end{tabular}
\end{table*}
\FloatBarrier

\end{document}